%% file: main.tex
\documentclass[pageno]{asplos/jpaper_arxiv}

\usepackage{balance} 
\usepackage{cleveref}
\usepackage{subcaption}

\newcommand{\asplossubmissionnumber}{278}

\usepackage[normalem]{ulem}

\begin{document}

\title{An Analysis of Collocation on GPUs for Deep Learning Training}

\author{
Ties Robroek\\
IT University of Copenhagen\\
titr@itu.dk
\and
Ehsan Yousefzadeh-Asl-Miandoab\\
IT University of Copenhagen\\
ehyo@itu.dk
\and
Pınar Tözün\\
IT University of Copenhagen\\
pito@itu.dk}


\date{}
\maketitle

\thispagestyle{empty}

\input{sections/0-abstract}
\input{sections/1-introduction}
\input{sections/2-background}
\input{sections/3-methodology}
\input{sections/4-results}

\input{sections/5-summary}

\input{sections/6-discussion}
\input{sections/7-conclusion}
\balance
\bibliographystyle{plain}
\bibliography{biblio}

\end{document}

%% file: sections/0-abstract.tex
\begin{abstract}
Deep learning training is an expensive process that extensively uses GPUs, but not all model training saturates modern powerful GPUs. \textit{Multi-Instance GPU (MIG)} is a new technology introduced by NVIDIA that can partition a GPU to better-fit workloads that do not require all the memory and compute resources of a full GPU. In this paper, we examine the performance of a MIG-enabled A100 GPU under deep learning workloads containing various sizes and combinations of models.
We contrast the benefits of MIG to older workload collocation methods on GPUs: \textit{naïvely} submitting multiple processes on the same GPU and utilizing \textit{Multi-Process Service (MPS)}.
Our results demonstrate that collocating multiple model training runs may yield significant benefits.
In certain cases, it can lead up to four times training throughput despite increased epoch time.
On the other hand, the aggregate memory footprint and compute needs of the models trained in parallel
must fit the available memory and compute resources of the GPU. 
MIG can be beneficial thanks to its interference-free partitioning, especially when the sizes of the models align with the MIG partitioning options. MIG's rigid partitioning, however, may create sub-optimal GPU utilization for more dynamic mixed workloads.
In general, we recommend MPS as the best performing and most flexible form of collocation for model training for a single user submitting training jobs.
\end{abstract}

%% file: sections/1-introduction.tex
\section{Introduction}
\label{sec:intro}

Today's GPUs are significantly more powerful than those of a decade ago. 
Modern GPUs, together with larger datasets, facilitate the exponential growth of deep learning models.
Many data scientists, however, do not require large models in practice.
For example, a problem may not have a large enough dataset to warrant a large model or the
ideal batch size for training the model may not be large enough
to utilize all of the GPU resources \cite{baunsgaardWT20, crossbow, hfta}.
This poses an hardware under-utilization issue
when training neural networks as the training process usually takes exclusive access to a GPU.
This problem gets exacerbated with each new GPU generation offering more hardware resources.

\textit{Workload collocation}
is a method for increasing hardware utilization
by running multiple applications at the same time over the same hardware resources.
When a workload does not require all of the resources available on a device,
a workload with additional applications can be considered. 
The result is that the device and its resources are shared among the collocated applications.
While workload collocation is heavily studied for CPUs \cite{DelimitrouK14, dicer, mesos},
its opportunities and challenges have been largely unexplored for modern GPUs. 
In addition, unlike CPUs, GPUs lack sophisticated resource-sharing methods such as virtual memory and fine-grained sharing.

Today, there are several methods for workload collocation on a GPU.
Firstly, multiple processes can be assigned to the same GPU simultaneously without any explicit process management.
Alternatively, the collocation can be more precisely managed, for example via NVIDIA's \textit{Multi-Process Service (MPS)}.
Finally, the latest generations of NVIDIA GPUs can be partitioned into fully isolated GPU instances at the hardware level via \textit{Multi-Instance GPU (MIG)}.

\begin{figure*}[th]
  \centering
  \includegraphics[width=\linewidth, trim={0.5cm 0.65cm 0.5cm 0.5cm},clip]{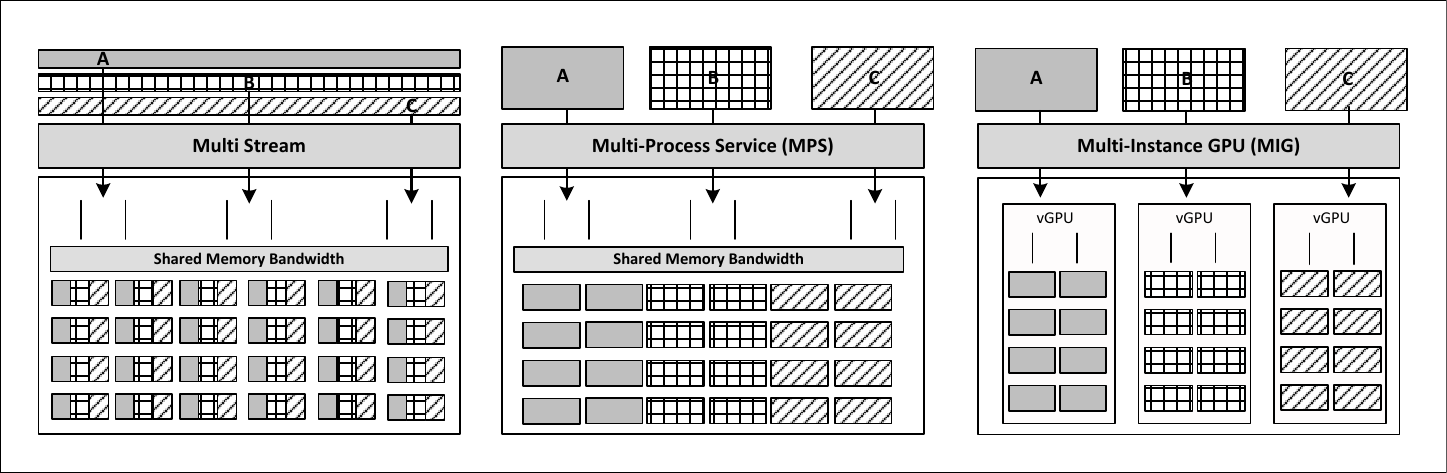}
  \vspace{-5mm}
  \caption{Collocation methods on modern NVIDIA GPUs. A, B, and C are different processes launched by the host CPU to run on the same GPU in a collocated manner using naïve approach (left-hand side), MPS (middle), and MIG (right-hand side).}
  \label{fig:gpu-collocation}
\end{figure*}

This paper analyzes different ways of collocating deep learning model training on NVIDIA GPUs.
Specifically, we investigate the strengths and limitations of the new MIG technology in contrast to the older methods.
We characterize the performance of the above-mentioned collocation methods on an A100 GPU.
We diversify our workload by considering three datasets (ImageNet, ImageNet64x64, Cifar10) representing different sizes (large, medium, small).
Furthermore, we acknowledge that the current deep learning landscape employs a wide variety of model architectures.
We investigate two popular convolutional models (ResNet, EfficientNetv2) and one transformer model (CaiT).
Additionally, we collocate a recommender model with a vision model to demonstrate the merits of workloads containing models that stress different parts of the hardware.
Our results demonstrate that:

\begin{list}{\labelitemi}{\leftmargin=1.5em}
    \item{When model training is unable to utilize the full GPU on its own,
    i.e., when running on our small- and medium-sized training cases, or when running cases that stress different parts of the GPU,
    training multiple models in collocated fashion presents considerable benefits.
    On the other hand, for large model training, 
    collocation provides either limited improvements to throughput as the GPU becomes over-saturated
    or cause model training to crash when the available GPU memory is not big enough to hold the combined memory footprint of the collocated models.
    }
    \item{On all the combinations we evaluated, MPS performs better than naïve and MIG collocation, achieving up to 80\% and 40\% higher throughput, respectively. It is also incredibly flexible, allowing single-user workloads to get the most out of the hardware with minimal setup required.}
    \item{MIG offers strict separation of the GPU's memory and compute resources across the collocated workloads, eliminating interference.
    It also allows multi-user collocation, unlike MPS.
    On the other hand, MIG-based collocation is more rigid, since MIG requires creating hardware partitions a priori. For the cases of well-defined workloads, one can create the ideal MIG partitions and leverage MIG-based collocation. However, for more dynamic workloads where the workload mix changes over time, MIG would require re-partitioning to perform well, whereas other collocation methods can still provide benefits.} 
\end{list}

The rest of the paper is organized as follows.
Firstly, \Cref{sec:background} gives background on GPU collocation techniques and surveys related work.
Then, \Cref{sec:methodology} describes our experimental methodology and setup,
and \Cref{sec:results} presents the results.
\Cref{sec:summary} outlines guidelines for collocation based on our results and touches on some challenges we encountered, and finally, \Cref{sec:conclusion} concludes the paper.


%% file: sections/2-background.tex
\section{Background}
\label{sec:background}
This section first provides background on different methods of collocation.
Then, we survey related work on workload collocation for deep learning.

\subsection{Collocation on GPUs}
\label{sec:collocation}
\Cref{fig:gpu-collocation} illustrates the three collocation methods we study in this paper.
We describe each of them briefly here.

\subsubsection{Naïve (or Multi-Stream)}
\label{sec:collocation:streams}
With CUDA 7, the option of running multiple processes at the same time using their own CUDA stream on the same GPU is introduced.
A \textit{CUDA stream} \cite{CUDA_streams} is a sequence of operations that execute on the GPU
(i.e., kernels and data transfers)
in the order in which they are issued. 
While operations within a stream are guaranteed to execute in the prescribed order,
operations in different streams can run concurrently.
This concurrency greatly helps with overlapping the data transfers between the host CPU and GPU.

We call this type of workload collocation the \textit{naïve} method 
since it offers a limited way for sharing GPU resources.
This is because the streams have to share the GPU compute resources in a time-based manner rather than having resources explicitly dedicated for each stream (\Cref{fig:gpu-collocation} left-hand side).

\subsubsection{Multi-Process Service (MPS)}
\label{sec:collocation:mps}

The \textit{multi-process service (MPS)} \cite{MPS-wp}
enables the host CPU to launch multiple processes on a single GPU.
Similar to naïve collocation, 
these processes share the GPU memory and memory bandwidth.
However, unlike naïve collocation,
the streaming multiprocessors (SMs) of the GPU are split across the different processes.
Assignment of the SMs is done by the MPS daemon automatically, unless explicitly stated by the user,
based on the provisioning of the GPU compute resources needed for each process
(\Cref{fig:gpu-collocation} middle).
While this provisioning introduces some process management overhead,
splitting resources this way avoids context switching of kernels from different processes on the same SM.
This reduces interference across the different processes compared to the naïve approach.
One limitation of MPS is that the processes have to be launched by a single user for security reasons.
Therefore, MPS cannot be used to collocate applications launched by different user accounts.

\subsubsection{Multi-Instance GPU (MIG)}
\label{sec:collocation:mig}
\textit{Multi-instance GPU (MIG)} \cite{mig-programming-guide} is the most recent collocation technology introduced with NVIDIA's Ampere GPUs.
It provides hardware support for splitting a GPU into smaller GPU instances of varying sizes. These GPU instances may run different processes each allowing these processes to run in parallel on the same GPU (\Cref{fig:gpu-collocation} right-hand side).

MIG-capable A100 GPUs consist of multiple slices. The memory of the GPU is split into 8 memory slices and the compute side is split into 7 compute slices, plus one reduced slice for the partition management overhead.
These can be combined into GPU instances providing a partitioning of the GPU.
A limitation of enabling MIG is that it does not allow for one model to be trained on multiple GPUs or GPU instances.
On the other hand, each partition is strictly separated in terms of hardware resources preventing any form of interference across partitions.

\begin{figure}[t]
  \centering
  \includegraphics[width=\linewidth, trim={0.5cm, 0.5cm, 0.5cm, 0.5cm}, clip]{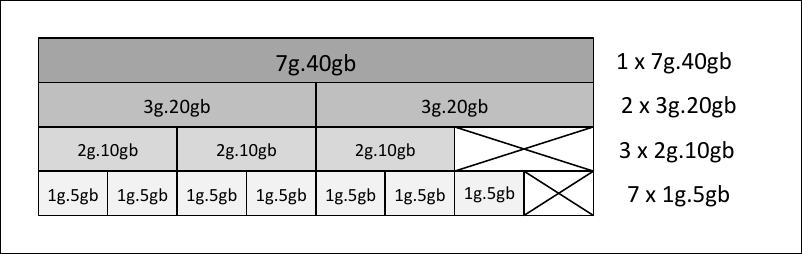}
  \vspace{-5mm}
  \caption{Possible MIG partitioning schemes on a NVIDIA A100-40GB GPU. Horizontals can overlap (collocation) but verticals cannot. For example, having a 3g.20gb instance is not compatible with 5x 1g.5gb instances (figure adapted from \cite{mig-programming-guide}).}
  \label{fig:MIG}
\end{figure}

An A100 GPU with 40GB memory supports several available partitioning profiles (see \Cref{fig:MIG}).
The smallest possible GPU instance is one with just one memory slice and one compute slice, \verb|1g.5gb|,
with 14 streaming multiprocessors (SMs) and 5GB of memory.
Consecutively, a \verb|2g.10gb| profile consists of two compute slices (28 SMs) and two memory slices (10 GB of memory).
The other available profiles are \verb|3g.20gb|, \verb|4g.20gb|, and \verb|7g.40gb|.
The last profile consists of almost all of the GPU resources.
However, using the GPU without MIG mode is not analogous to running this large profile as the compute capability of the GPU is hampered slightly due to MIG management overhead; i.e. the reduced compute slice as mentioned above (10 SMs).
The ideal GPU instance sizes may vary workload to workload following the memory and compute needs of the models.

Many different partitions are possible as long as the maximum resource capacity is not exceeded.
For example, splitting the GPU into a \verb|4g.20gb| and \verb|1g.5gb| instance is possible
but two \verb|4g.20gb| instances would exceed the compute resources of the device.
There is, however, a notable exception.
While a split of one \verb|4g.20gb|, one \verb|2g.10gb|, and one \verb|1g.5gb| instance is possible, one cannot proceed with a split of one \verb|4g.20gb| and one \verb|3g.20gb| instance, despite the values appearing to sum up to the maximum resources of the device. 
Such partitioning rules are set by the GPU itself,
and the allowed set of instances and configurations varies across different types of Ampere GPUs (A100, A30)
as well as the NVIDIA GPU architectures that come after (e.g., H100).

Finally, the amount of memory slices and the amount of compute slices may differ in a partitioning.
Specifically, a GPU instance may be split into multiple compute instances from the compute side with unified memory.
This can be useful when compute requirements and memory requirements do not follow the same pattern. For example, one could run a memory intensive model and a compute intensive model with isolated compute instances on a single GPU instance.

\subsection{Related work}
\label{sec:related}
Weng et al. \cite{studyGPU} observe 6000 GPUs on Alibaba clusters and highlight the challenges of cluster scheduling mechanisms that result in dramatically low GPU utilization.
Jeon et al. \cite{234916} perform a similar experimental analysis focusing on deep learning training on a Microsoft GPU cluster.
These works motivate studying workload collocation on GPUs as a way to overcome such low utilization. 

Proposals for collocation mechanisms on GPUs have been studied in two dimensions: software-based approaches and hardware-based ones.

Software-based approaches either focus on developing better primitives for collocation on GPUs or provisioning the resources of GPUs for running multiple applications. 
cuMAS \cite{cuMAS} is a host-side CUDA task scheduler, which receives multiple CUDA calls and reorders them based on data transfer behavior to increase overall system utilization.
Ravi et al. \cite{2011-framework-gpu-sharing} propose a framework as a transparent layer for executing applications within virtual machines to share one or more GPUs.
Horus \cite{Horus} uses machine learning for predicting GPU utilization of deep learning training tasks. Afterwards, it feeds the cluster scheduler and resource manager with the information to make better decisions for collocating different workloads.

In contrast, hardware approaches propose micro-architectural changes to GPUs to enable finer-grained and more precise multi-application execution within a GPU considering performance, utilization, and quality of service trade-offs \cite{10.1145/3508036, 8327010, 7446078, 8192477, 7551396, 10.1145/3205289.3205311}.

MIG is a relatively new technology and there have not been many works that thoroughly explore its possibilities.
HFTA \cite{hfta} is a mechanism to fuse multiple model training runs for hyper-parameter tuning into one training run. The authors show the effectiveness of HFTA compared to using MPS or MIG to run multiple training runs in parallel.
MISO \cite{li2022using} runs MPS on a \verb|7g.40gb| MIG instance to predict the best MIG configuration for different jobs. 

Finally, similar to our work, Li et al. \cite{9835424} characterize performance of MIG using deep learning models focusing on time and energy metrics. Their methodology is different than and complementary to ours. It covers a variety of deep learning use cases, but doesn't consider sizing the models up and down. Furthermore, they don't compare against other forms of collocation such as MPS. 

In general, our work is complementary to these works since we focus on an experimental methodology to investigate the strengths and limitations of MIG in contrast to the older collocation techniques such as MPS and naïve collocation and by using workloads of different sizes.

%% file: sections/3-methodology.tex
\section{Setup \& Methodology}
\label{sec:methodology}
This section details our experimental setup and methodology.
First, \Cref{sec:system} introduces the hardware system used for conducting our experiments.
Next, \Cref{sec:metrics} defines our metrics, their relevance, and how we measure them.
\Cref{sec:datasets_models} describes the models and datasets used in this study. 
Finally, \Cref{sec:experiments} describes the list of the experiments 
and the software framework used to run the experiments.

\subsection{System}
\label{sec:system}
Our experiments run on a DGX Station A100,
composed of an AMD EPYC 7742 CPU and four A100 40GB GPUs.
The system is a pre-packaged solution provided by NVIDIA running \textit{DGX OS}, a variant of \textit{Ubuntu 20.04.4 LTS}.
The CPU consists of 64 cores, 128 threads, operating at a base clock of 2.25 GHz with a boost clock of 3.4GHz \cite{amd-epyc-7742-product}. 
There is 256MB of L3 cache and 512GB of DRAM available.
Each of the  A100 GPUs have 40GB of VRAM and support up to 7 MIG instances with at least 5 GB of memory per instance (see \Cref{sec:collocation:mig}).

\subsection{Metrics}
\label{sec:metrics}
The goal of this paper is to investigate the performance of different GPU collocation techniques
instead of improving the accuracy of models for a particular use case.
Therefore, the set of metrics we focus on are related to how the model training interacts with and gets impacted by the GPU resources and the management of these resources.

\textbf{Time per epoch} 
is the time it takes to finish a single epoch of training for a particular model.
GPUs are used in deep learning in order to reduce training time by exploiting the embarrassingly parallel nature of most deep learning computations.
Therefore, time per epoch is the most fundamental metric to look at in our study.
We time the second epoch of training, skipping the first one as a warm-up epoch.

\textbf{GPU utilization}
depicts how much the GPU is being used.
We are interested in how active the whole GPU is depending on the workload executed and collocation mechanism used. 

We use \textbf{SMACT} (SM Activity) and \textbf{SMOCC} (SM Occupancy) to track GPU utilization.
SMACT
is the fraction of active time on an SM, averaged over all SMs. 
SMOCC
is the degree of parallelism within an SM relative to the maximum degree of parallelism supported by the SM.
The combination of these two metrics provides a better indication of whether the GPU is in use than other available metrics \cite{yousefzadeh2023profiling}.
These metrics are reported by the Data Center GPU Manager (\texttt{dcgm}) \cite{dcgm-user-guide}.
They can be tracked for the whole GPU when there are no MIG instances available and per MIG instance when there are. We aggregate SMACT and SMOCC for MIG to compare readings from MIG instances to those from the entire GPU. 

\textbf{Memory footprint}
is the total memory space (in GB) allocated by all of the collocated models on the GPU.
This metric is especially crucial for reasoning about the failed collocation attempts.
Specifically, we measure the memory requirement after a full epoch of training to signify how much memory is needed for the model to train.
We use \texttt{nvidia-smi} to collect the memory consumption for the whole GPU.

As naïve collocation and MPS share memory across all the runs within the workload, the available GPU memory must be able to accommodate the memory footprint for the collocated runs (\Cref{sec:collocation}).
If the sum of memory required exceeds the capacity of the device, the workload will fail due to running out of memory.
For MIG (\Cref{sec:collocation:mig}), the memory available per GPU instance needs to be large enough to accommodate the memory footprint of the models mapped to that instance.



\subsection{Models \& Datasets}
\label{sec:datasets_models}

\begin{table}[]
\centering
\vspace{-3mm}
\caption{Models \& Datasets}
\resizebox{8.1 cm}{!} {
\begin{tabular}{|l|c|c|c|}
\hline
Model & Dataset & \#Parameters & Size \\ \hline \hline 
ResNet26 & Cifar10  & 17M & small \\ \hline 
ResNet50 & ImageNet64 & 24M & medium \\ \hline 
ResNet152 & ImageNet & 59M & large \\ \hline \hline 
EfficientNet\_v2\_s & Cifar10 & 22M & small  \\ \hline 
EfficientNet\_v2\_s & ImageNet64 & 22M & medium \\ \hline 
EfficientNet\_v2\_s & ImageNet  & 22M & large \\ \hline \hline 
CaiT\_xxs24\_224 & ImageNet  & 12M & large \\ \hline \hline 
DLRM & Criteo Terabyte  & 24B & very large \\ \hline 
\end{tabular}
}
\label{table:datasets}
\end{table}

Deep learning achieves state-of-the-art models for a variety of use cases, such as speech recognition, image classification, and sentiment analysis.
These fields feature a plethora of models with different compute and memory requirements.
\Cref{table:datasets} lists the models and datasets used in our experiments.

\subsubsection{Models}
\label{sec:models}
We select three model architectures representing a large range of image classification models in addition to a recommender model.
In terms of the interaction with the hardware, deep learning applications that leverage GPUs are either compute- or memory-intensive on the GPU. This includes other models such as for speech recognition and object detection.
The type of interaction with the hardware is the determining factor for the behavior of the collocated training runs.
By including a variety of compute-intensive neural network architectures under image classification and the memory-intensive recommender model, this paper aims at covering a good representative set of training cases.

\textbf{ResNet} is a deep convolutional neural network that has been around since 2016 \cite{resnet, resnetv2}.
In addition to being a popular choice for image classification and segmentation, 
ResNets \cite{resnetsurvey2, resnetsurvey1} can be scaled up and down in size,
which makes them ideal for benchmarking over varying hardware resources.
This helps with creating workloads of varying sizes for testing the different workload collocation options,
especially the different MIG partitions.
We train ResNet26, ResNet50, and ResNet152 models to create \textit{small}, \textit{medium}, and \textit{large} workloads, respectively.
The medium model has significantly more parameters compared to the small one,
and the large model has about twice the parameters of the medium model.
We train these models on datasets corresponding to their size (see \Cref{sec:data} for details).

\textbf{EfficientNetv2} is a recent convolutional neural network \cite{DBLP:journals/corr/abs-2104-00298} for image classification.
EfficientNetv2's architecture is focused on delivering high performance while limiting the size of the model.
In order to satisfy memory constraints on a single GPU, we exclusively train the small version of EfficientNetv2.
We train the small version on all three image datasets creating \textit{small}, \textit{medium}, and \textit{large} workloads.

\textbf{CaiT} is a visual transformer model.
Unlike many other transformers for image classification, CaiT achieves high performance without the need of extra data \cite{DBLP:journals/corr/abs-2103-17239}. 
We use the smallest version of CaiT (\textit{xxs}) to satisfy memory constraints on the GPUs in our system (\Cref{sec:system}).
Since most transformer model architectures for image classification start at a relatively large size compared to their convolutional counterparts,
we only create a \textit{large} workload with CaiT, training it on the largest image dataset.

\textbf{DLRM} is a recommendation model. Unlike the previous vision models, this model is used to provide personalizations and recommendations based on past user behavior \cite{DLRM19}. The model is significantly less GPU compute-heavy than vision models but requires large amounts of CPU and GPU memory. We use the MLPerf configuration of the model as provided by the authors.
We train DLRM on the Criteo Terabyte dataset \cite{criteo}. Training DLRM for an epoch on this dataset takes significantly longer than training any of the vision models for an epoch. We therefore look at the rate at which DLRM goes through the data instead of the time that a full epoch takes ($\sim$ 4.3 days).

\subsubsection{Datasets}
\label{sec:data}
We accompany the vision models with three datasets of varying sizes, forming small, medium, and large workloads.
We also include a very large dataset for the recommender model.

For our \textbf{\textit{small}} image dataset we have CIFAR-10 \cite{Krizhevsky09learningmultiple} (163 MB),
containing 60,000 labeled $32\times32$ pixel images divided over 10 classes.
The dataset is split into 50,000 training images and 10,000 test images.

Our \textbf{\textit{medium}} image dataset is a downsampled version of the large dataset, ImageNet2012, called ImageNet64$\times$64 \cite{smallimagenet} (12 GB). We will refer to this dataset as ImageNet64.

For our \textbf{\textit{large}} image dataset, we use the unmodified ImageNet2012 \cite{ILSVRC15} (138 GB), referred to as ImageNet.
Imagenet2012 is a collection of 1,431,167 labeled images from 1,000 different classes.
The dataset is split into 1,281,167 training, 50,000 validation, and 100,000 test images.
Unlike CIFAR-10, the dataset is not balanced and the images are not all uniform in size.
Every picture is resized to $224 \times 224$ using the \textit{nearest pixel} interpolation method to conform with the size of images used in the original ResNet specification \cite{resnet}. 

Finally, we use the Criteo 1TB Click Logs dataset~\cite{criteo} for training the DLRM model. The dataset consists of online advertisement click-through logs and contains 24 days of data. Crucially, we run the model in \texttt{memory-map} mode. This pre-processes the data without loading all in CPU memory at once, preventing the system from running out of memory.

\begin{figure*}[!ht]
\centering
\begin{subfigure}{.329\textwidth}
  \centering
\includegraphics[width=\linewidth]{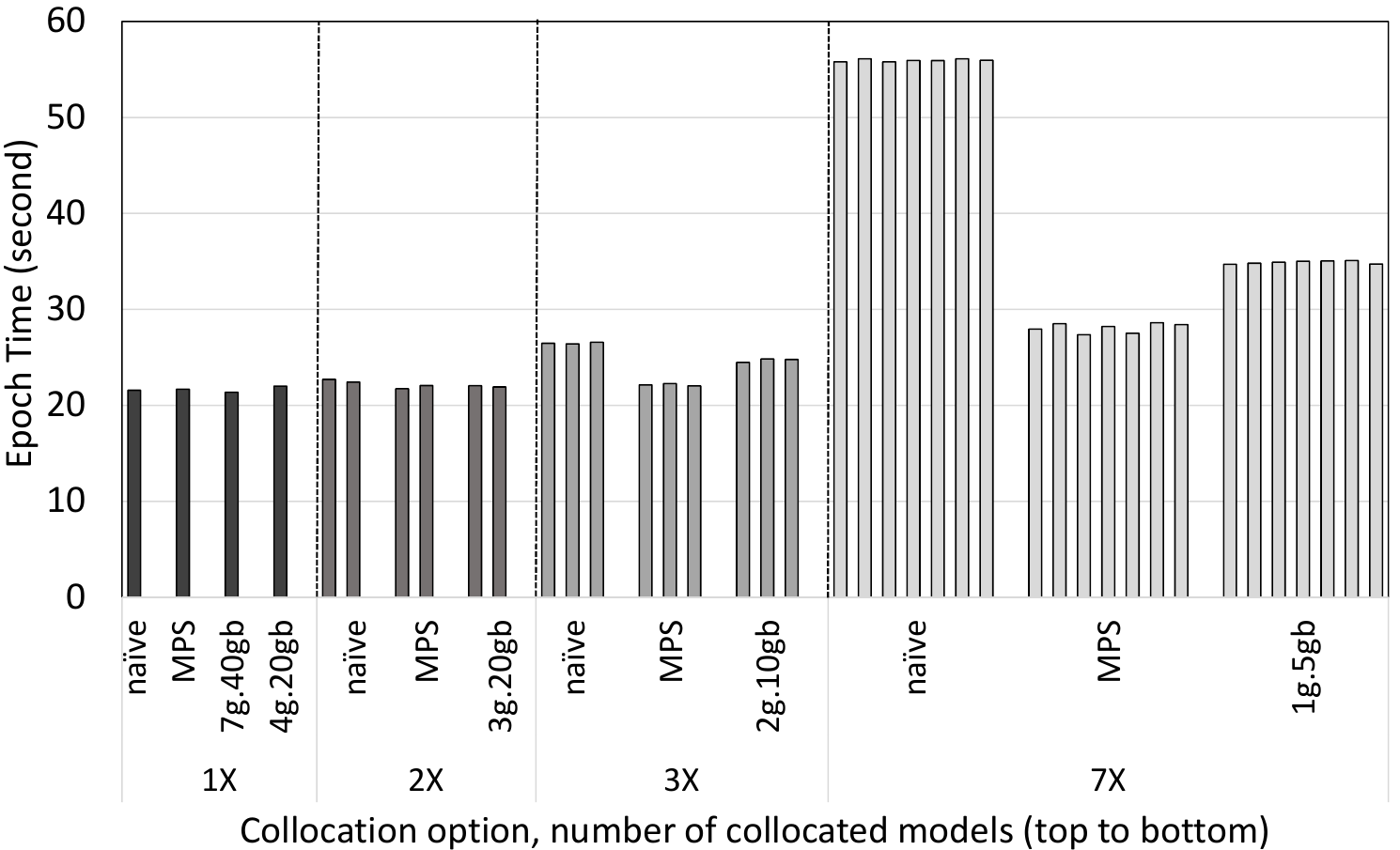}
\vspace{-5mm}
  \caption{Epoch time}
  \label{fig:time-resnet32-small}
\end{subfigure}
\begin{subfigure}{.329\textwidth}
  \centering
\includegraphics[width=\linewidth]{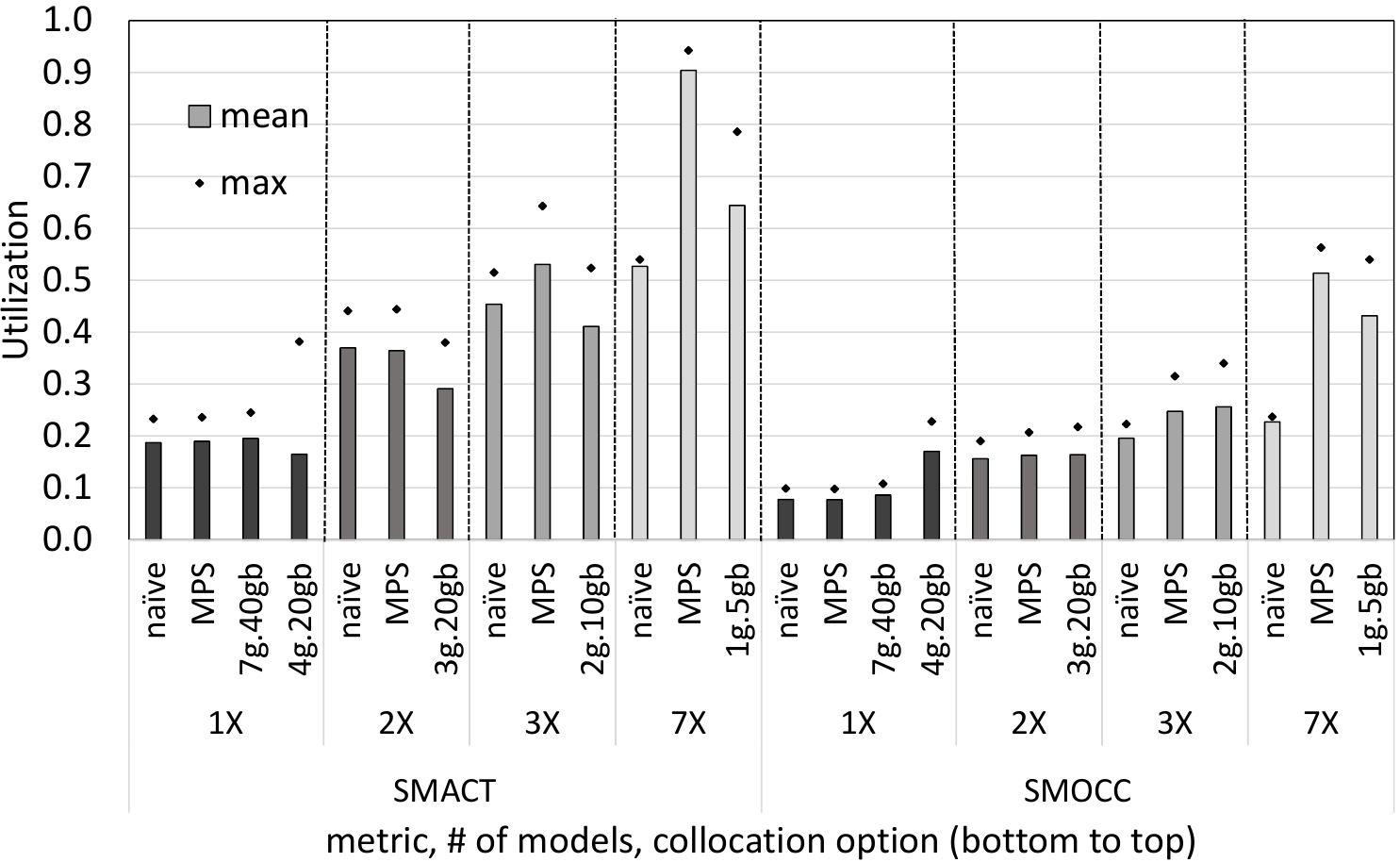}
\vspace{-5mm}
  \caption{GPU utilization}
  \label{fig:gract-resnet32-small}
\end{subfigure}
\begin{subfigure}{.329\textwidth}
  \centering
\includegraphics[width=\linewidth]{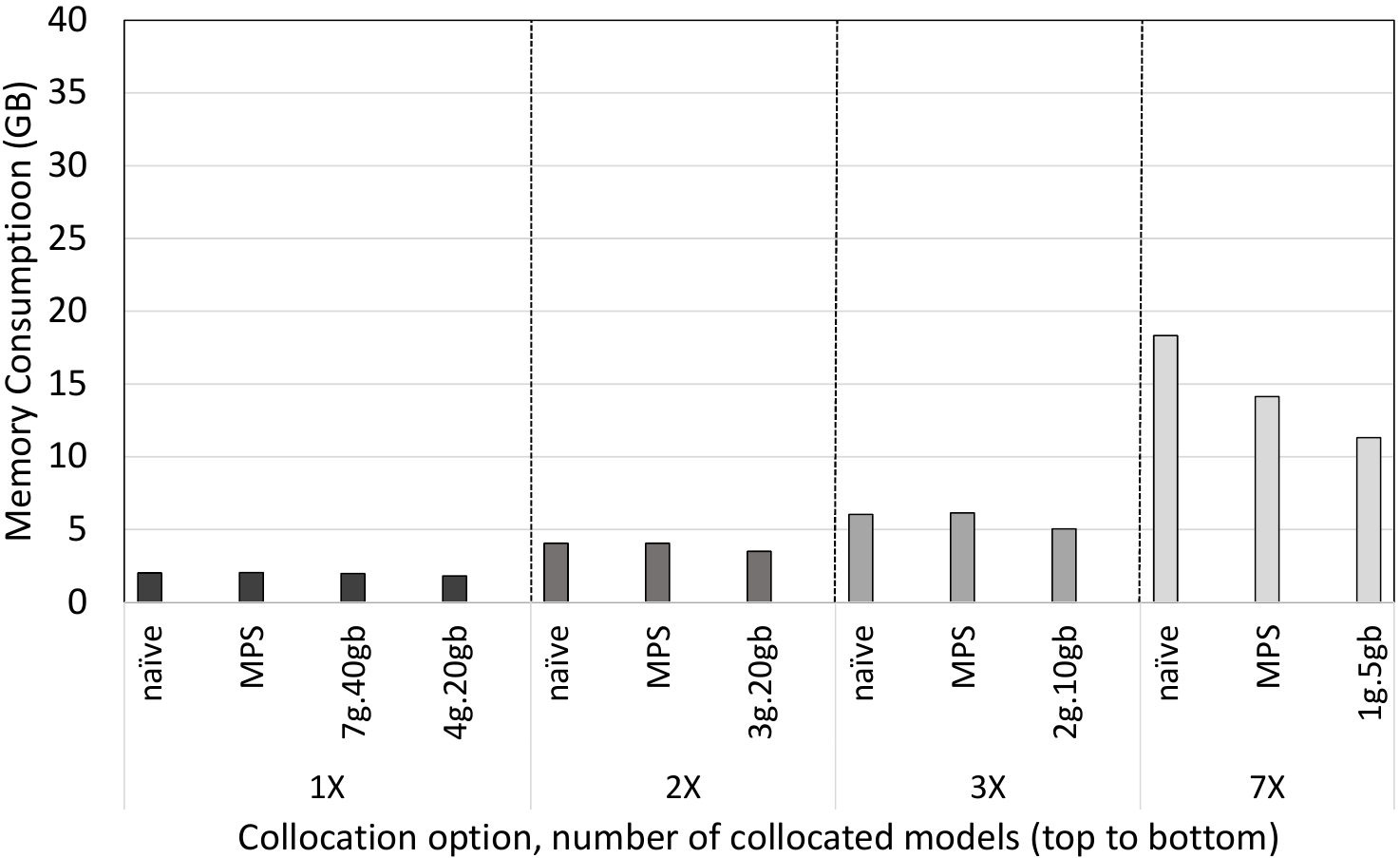}
\vspace{-5mm}
 \caption{Memory footprint}
  \label{fig:drama-resnet32-small}
\end{subfigure}
\vspace{-4mm}
\caption{Small: ResNet26 + Cifar10 (batch size = 32).}
\label{fig:small-resnet32-all}
\end{figure*}


\begin{figure*}[!ht]
\centering
\begin{subfigure}{.329\textwidth}
  \centering
\includegraphics[width=\linewidth]{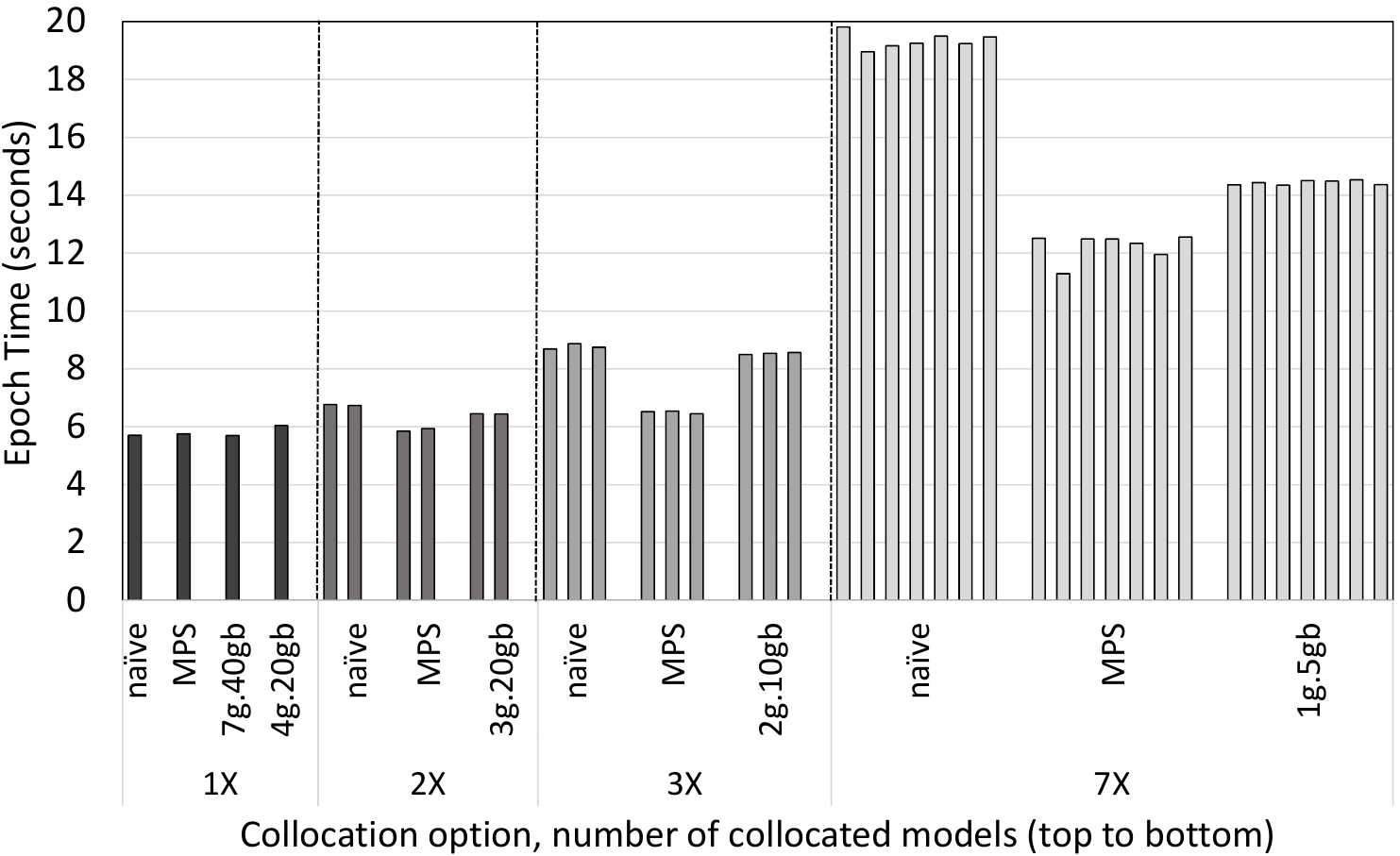}
\vspace{-5mm}
  \caption{Epoch time}
  \label{fig:time-resnet128-small}
\end{subfigure}
\begin{subfigure}{.329\textwidth}
  \centering
\includegraphics[width=\linewidth]{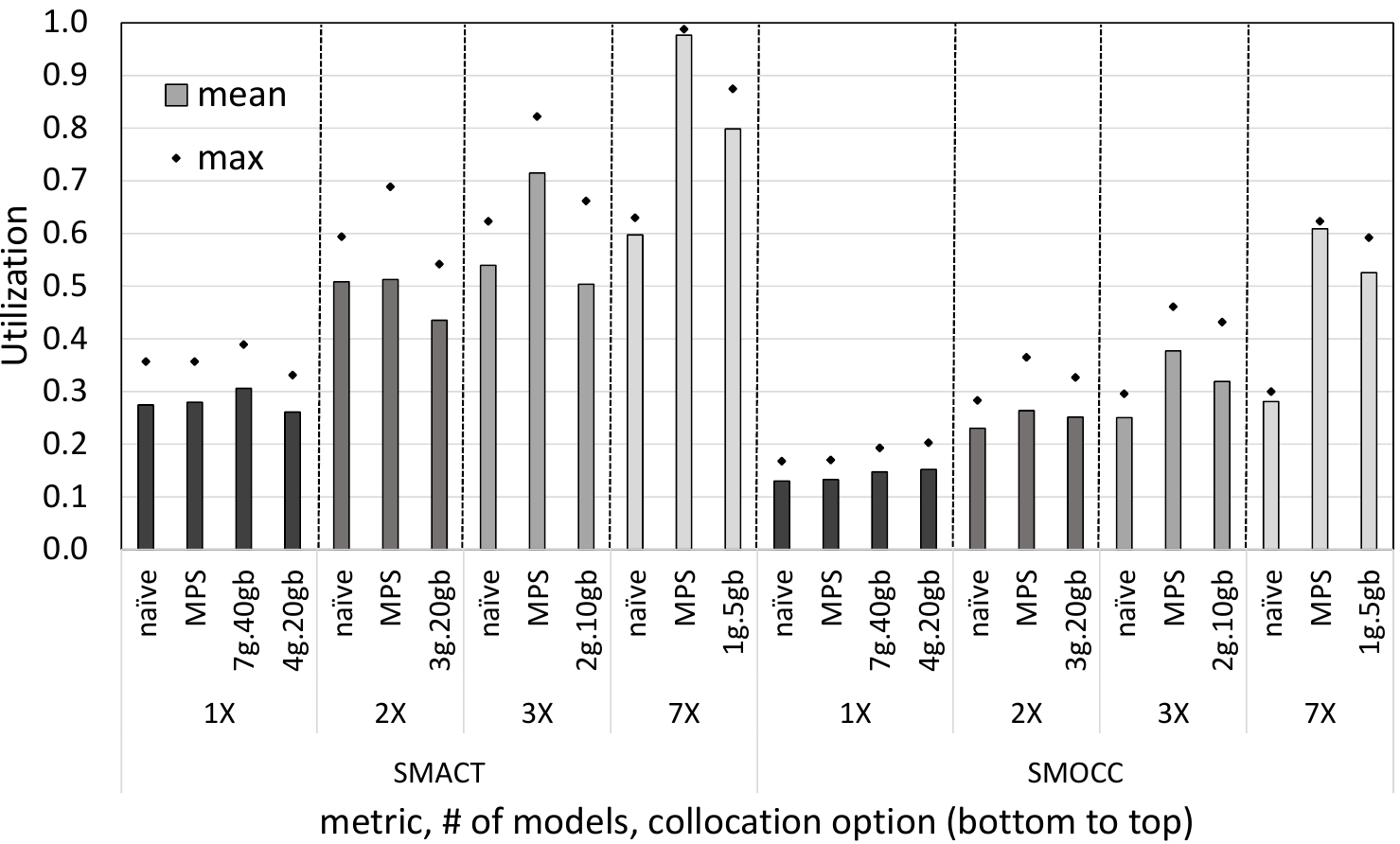}
\vspace{-5mm}
  \caption{GPU utilization}
  \label{fig:gract-resnet128-small}
\end{subfigure}
\begin{subfigure}{.329\textwidth}
  \centering
\includegraphics[width=\linewidth]{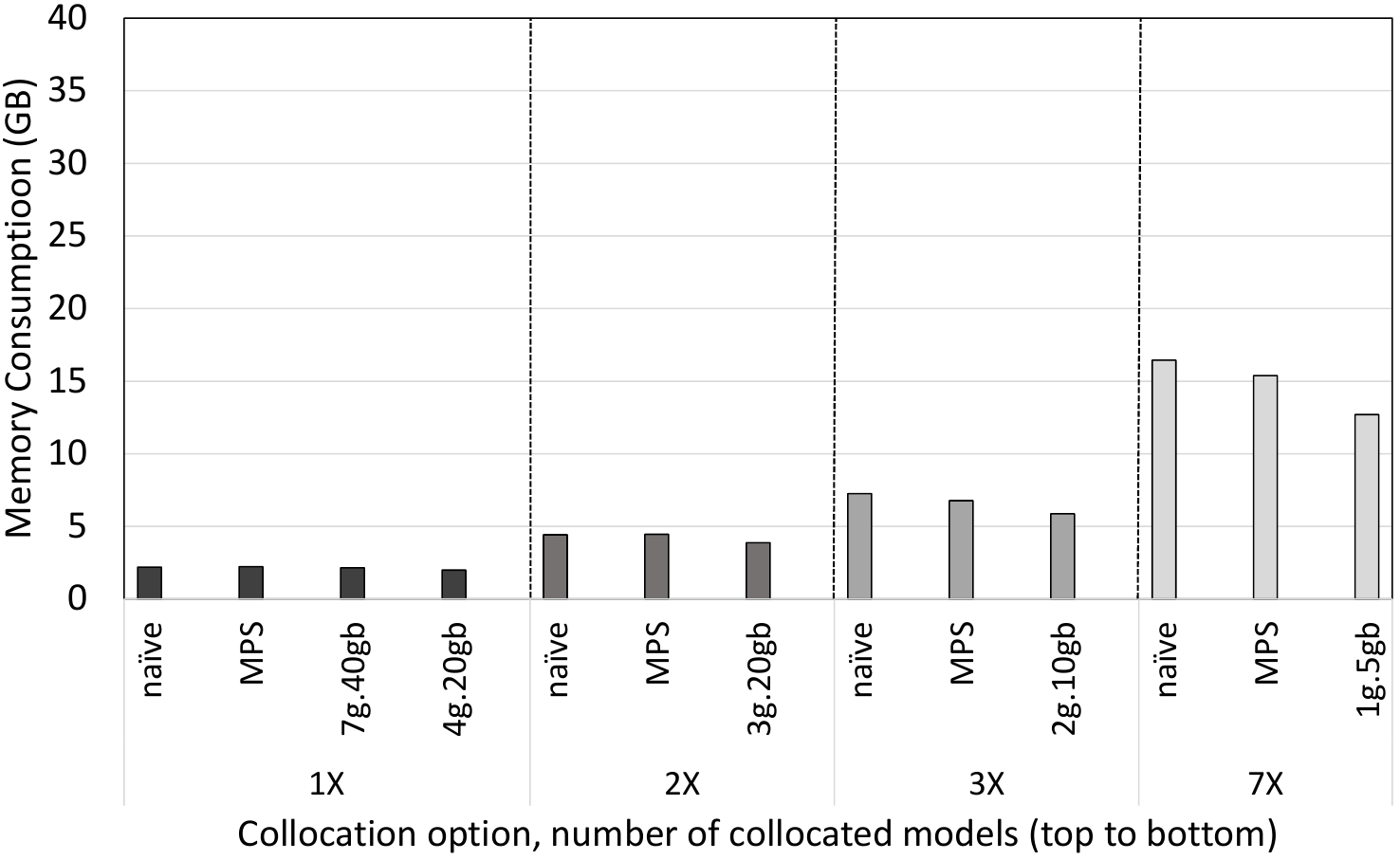}
\vspace{-5mm}
  \caption{Memory footprint}
  \label{fig:drama-resnet128-small}
\end{subfigure}
\vspace{-4mm}
\caption{Small: ResNet26 + Cifar10 (batch size = 128).}
\label{fig:small-resnet128-all}
\end{figure*}


\begin{figure*}
\begin{subfigure}{.329\textwidth}
  \centering
  \includegraphics[width=\linewidth]{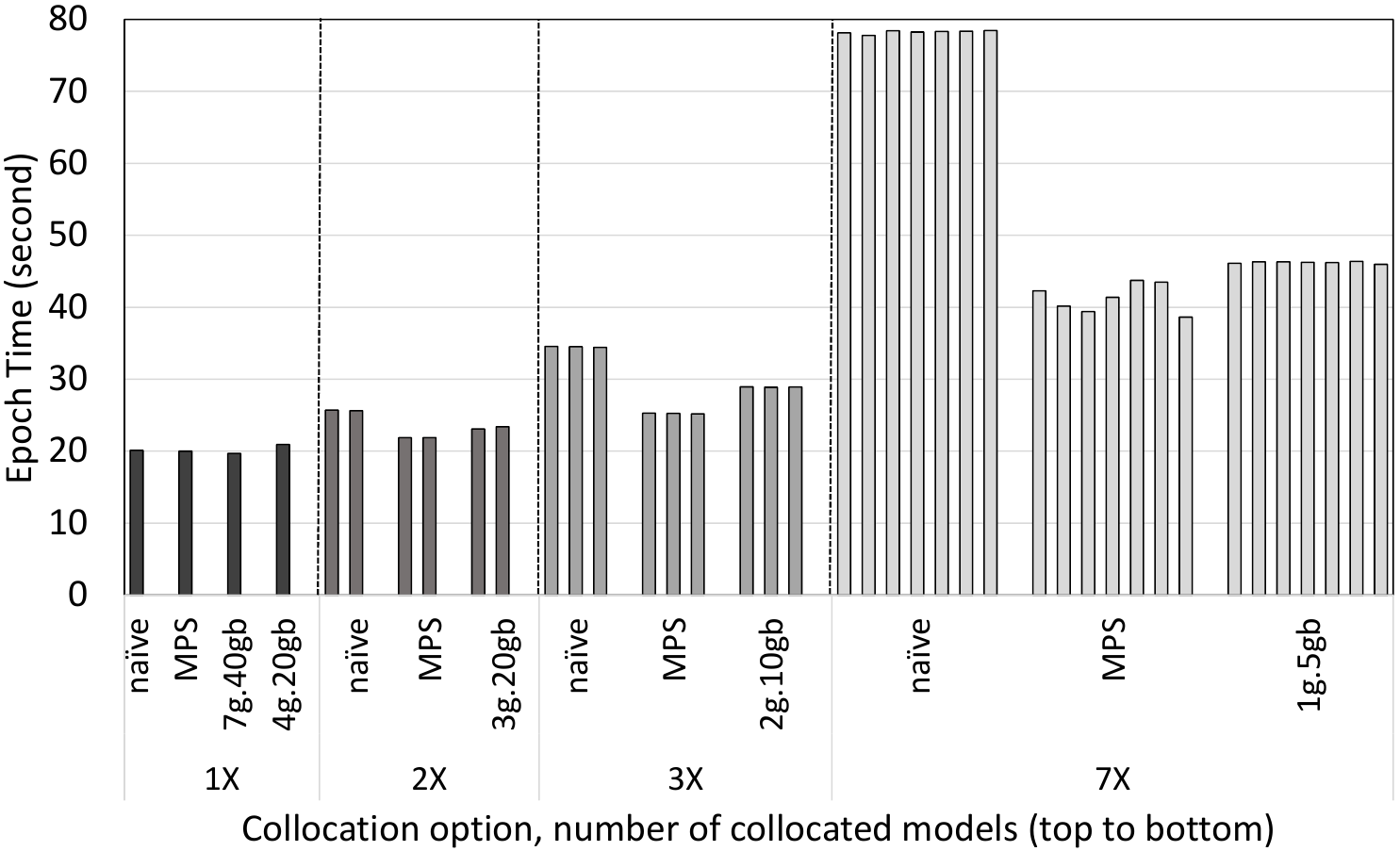}
  \caption{Epoch time}
  \label{fig:time-efnet-small}
\end{subfigure}
\begin{subfigure}{.329\textwidth}
  \centering
\includegraphics[width=\linewidth]{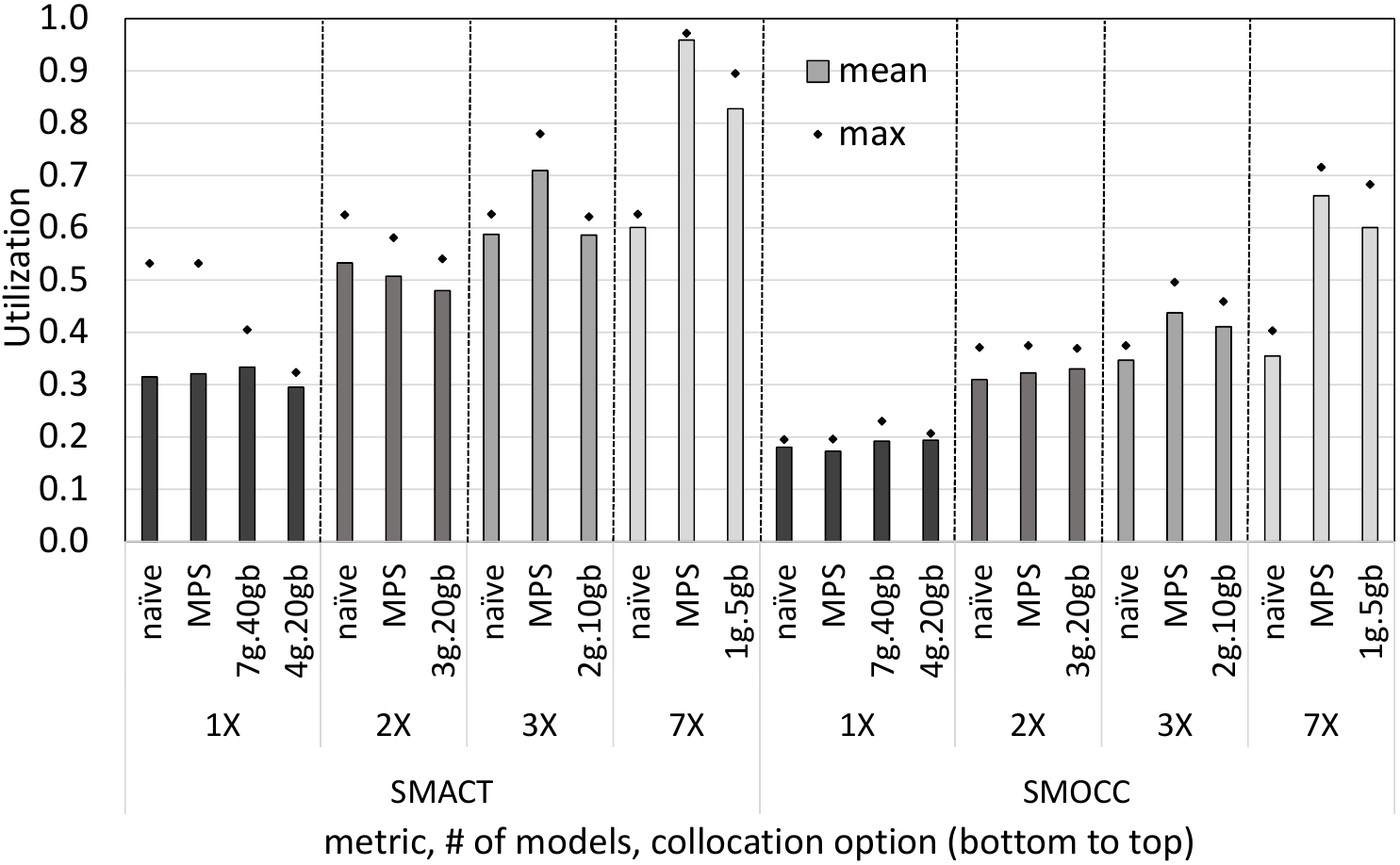}
  \caption{GPU utilization}
  \label{fig:gract-efnet-small}
\end{subfigure}
\begin{subfigure}{.329\textwidth}
  \centering
\includegraphics[width=\linewidth]{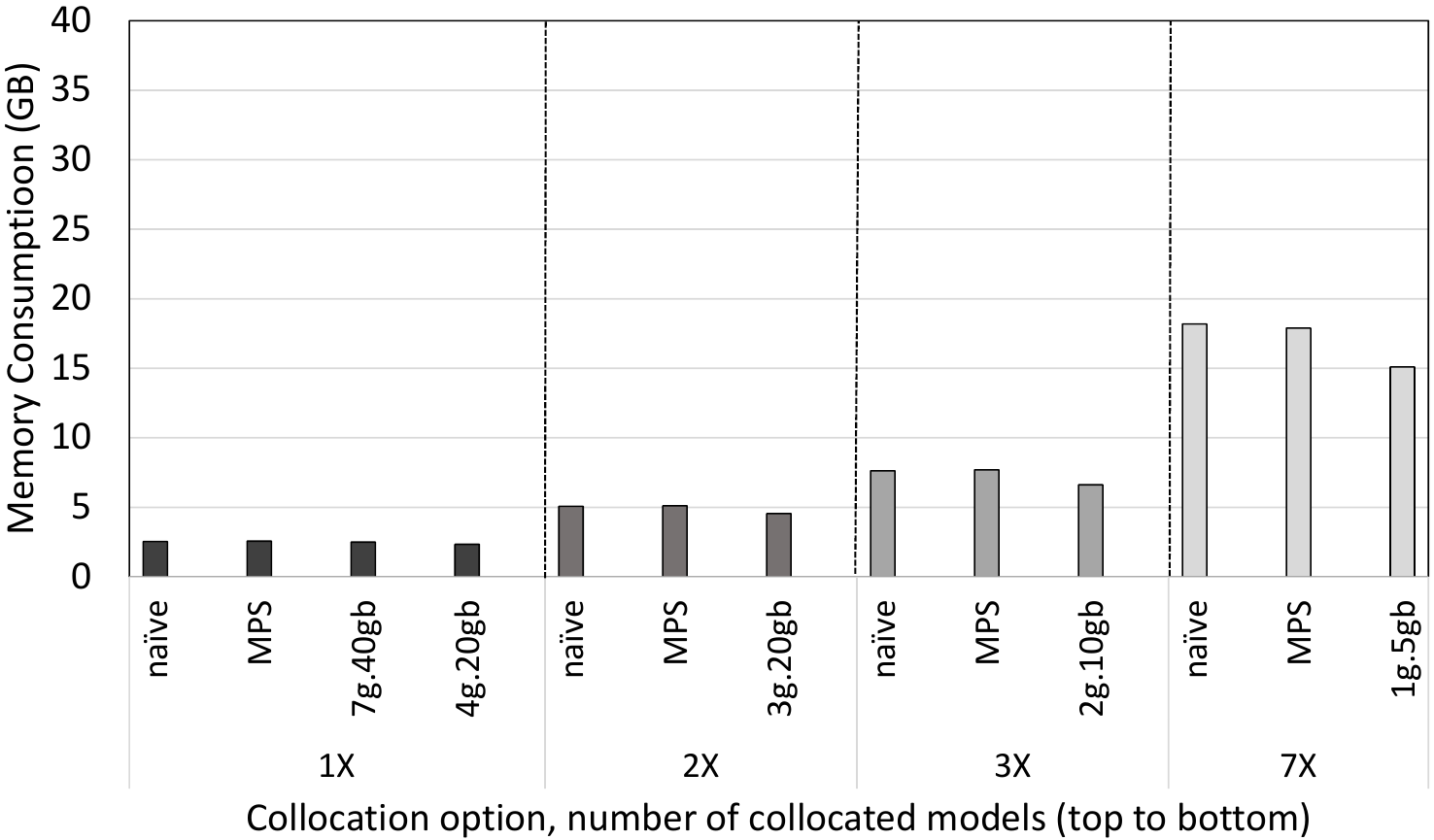}
  \caption{Memory footprint}
  \label{fig:drama-efnet-small}
\end{subfigure}
\vspace{-3mm}
\caption{Small: EfficientNet\_s + Cifar10 (batch size = 128).}
\label{fig:small-efnet-all}
\end{figure*}

\subsection{Experiments}
\label{sec:experiments}
We devise experiments with varying dataset sizes and models
to assess the performance of collocating deep learning training under different loads.

We orchestrate the execution of the workloads via a custom framework built on top of the machine learning platform MLflow \cite{mlflow_framework}.
This allows running the experiments in a systematic and controlled way.
We create workloads containing either a single or collocated model(s).
Workloads are automatically executed in sequence by the framework and in controlled conda environments.
Any setup requirements for MIG partitioning and the MPS daemon are performed before training.
The GPUs are cleaned after every workload, removing any previous configurations such as MIG partitions.
The framework comes with a set of listeners that automatically take measurements while the models are training.
In this case, every model has a 
DCGMI and NVIDIA-SMI listener bundled with it
to collect the metrics of interest (\Cref{sec:metrics}).
We source the vision models from the TIMM library \cite{rw2019timm} and the recommender model from Facebook Research \cite{DLRM19}, and are using the latest version of PyTorch as of writing (2.0) \cite{NEURIPS2019_9015}.

\textbf{Uniform collocation.}
We default to a batch size of 128 for most of our experiments.
We additionally train the \textit{ResNet} models on a batch size of 32 to observe the impact of batch size.
Based on our preliminary experiments with some of the large, hence longer-running, models and dataset,
we observed that the behavior of time-per-epoch, GPU utilization, and memory consumption (\Cref{sec:metrics}) does not drastically change from the second epoch on.
As the first epoch of the vision models tends to be slower than subsequent epochs, we let the vision models warm up for one epoch and report the measurements from the second epoch, providing representative information. 

We determine several model training collocation options following the available MIG profiles (\Cref{sec:collocation:mig}):
\begin{list}{\labelitemi}{\leftmargin=1.5em}
   \item One model: MIG 7g.40gb or 4g.20gb
   \item Two models: MIG 3g.20gb
   \item Three models: MIG 2g.10gb
   \item Seven models: MIG 1g.5gb
\end{list}

These are based on the maximum amount of instances that can be allocated at the same time for a given MIG profile.
We also create the corresponding collocation experiment for the non-MIG collocation methods (\Cref{sec:collocation}). 
For example, for the \verb|1g.5gb| profile, there can be a maximum of 7 MIG instances present on the GPU at the same time
allowing for 7 models to be trained in parallel, each model on a separate instance.
We contrast this setup with training 7 models in parallel using naïve collocation and MPS. 
These form our initial set of experiments collocating uniform training runs.

The \verb|7g.40gb| and \verb|4g.20gb| profiles do not allow for any parallel instances of the same size since there are not enough compute or memory left for such an instance.

Finally, the experiments with the full MIG profile, \verb|7g.40gb|, and MPS without collocation have the purpose to explore the performance impact of the respective technologies for a GPU being enabled, in comparison to a case where they are not.

\begin{table}[]
\centering
\caption{Mixed Vision Workloads}
\resizebox{7.1 cm}{!} {
\begin{tabular}{|c|c|c|c|}
\hline
           & Small (S) & Medium (M) & Large (L)     \\ \hline \hline
Model      & ResNet26 & ResNet50   & ResNet152 \\ \hline
Dataset    & CIFAR10  & ImageNet64 & ImageNet  \\ \hline
Batch size & 128      & 128        & 32        \\ \hline
Instance   & 1g.5gb   & 2g.10gb    & 4g.20gb   \\ \hline \hline
S+M        & 1x       & 1x         &      -     \\ \hline
S+S+M      & 2x       & 1x         &      -     \\ \hline
S+S+M+M    & 2x       & 2x         &      -     \\ \hline
S+S+S+M    & 3x       & 1x         &      -     \\ \hline
S+M+M+M    & 1x       & 3x         &      -     \\ \hline
S+L        & 1x       &        -   & 1x        \\ \hline
M+L        &     -     & 1x         & 1x        \\ \hline
\end{tabular}
}
\label{table:mixedvision}
\end{table}

\textbf{Mixed collocation.}
We also create collocation experiments with mixed sets of ResNet models and datasets,
based on the performance of the models in the prior experiments.
In order to satisfy the memory constraints, we train the small and medium models with batch size 128 and the large models with batch size 32. We limit the amount of collocated models to four in order to provide a clear scope that should cover most use-cases.
\Cref{table:mixedvision} lists all the mixed ResNet training collocation options we experiment with.

Similarly, we collocate the recommender model with a single large ResNet model with batch size 32.
Due to the size of the recommender model, we limit our experiment to 2-way collocation. As the memory requirements of the recommender exceed that of even the 20GB MIG partitions, we issue a \texttt{7g.40gb} MIG GPU instance and split this instance into two compute instances
(see \Cref{sec:collocation:mig}). This yields a \texttt{3c.7g.40gb} and a \texttt{4c.7g.40gb} compute instance to train the models on,
where the 40GB memory is shared between the collocated training runs.

%% file: sections/4-results.tex

\begin{figure*}[!ht]
\centering
\begin{subfigure}{.329\textwidth}
  \centering
\includegraphics[width=\linewidth]{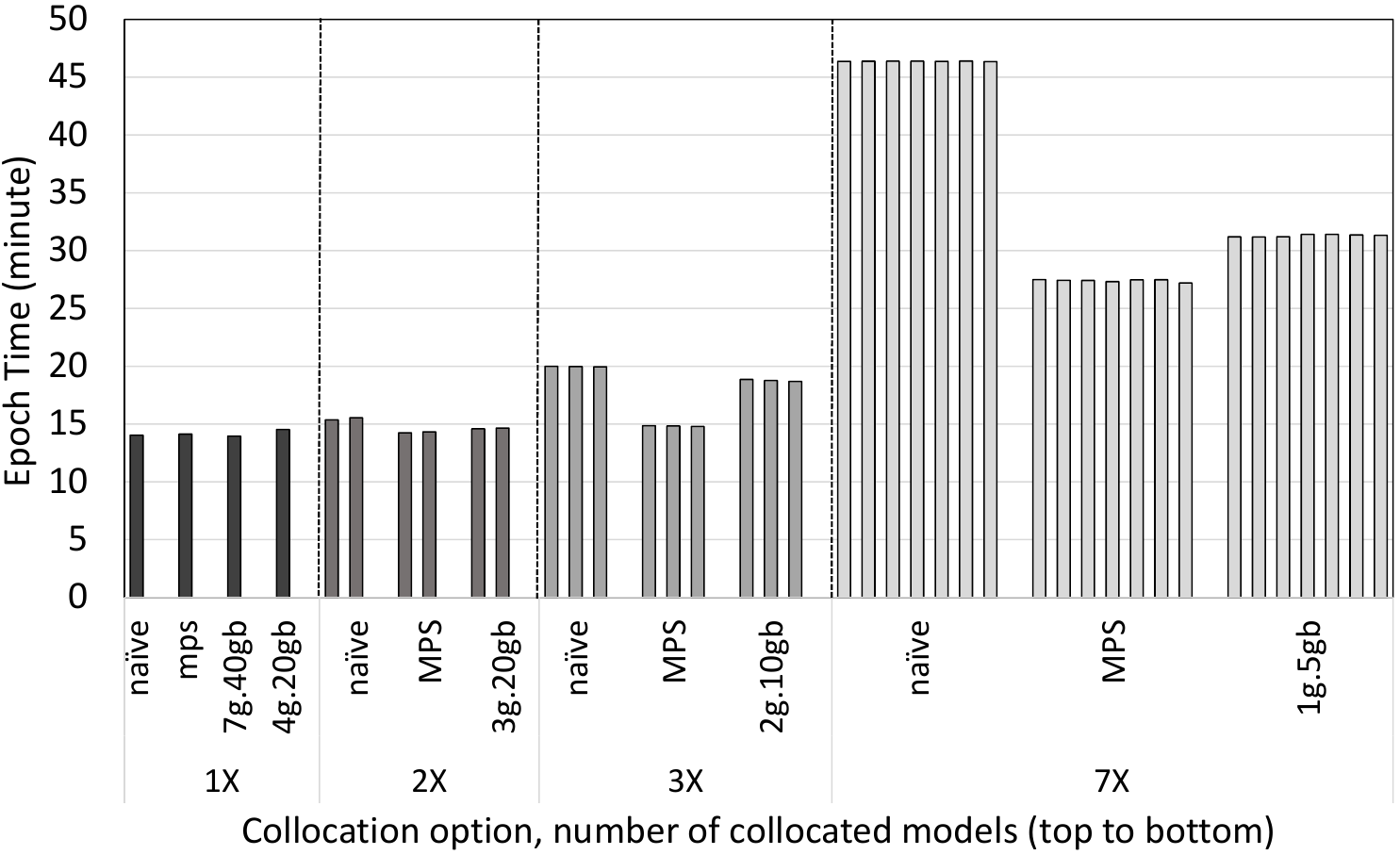}
\vspace{-5mm}
  \caption{Epoch time}
  \label{fig:time-resnet32-medium}
\end{subfigure}
\begin{subfigure}{.329\textwidth}
  \centering
\includegraphics[width=\linewidth]{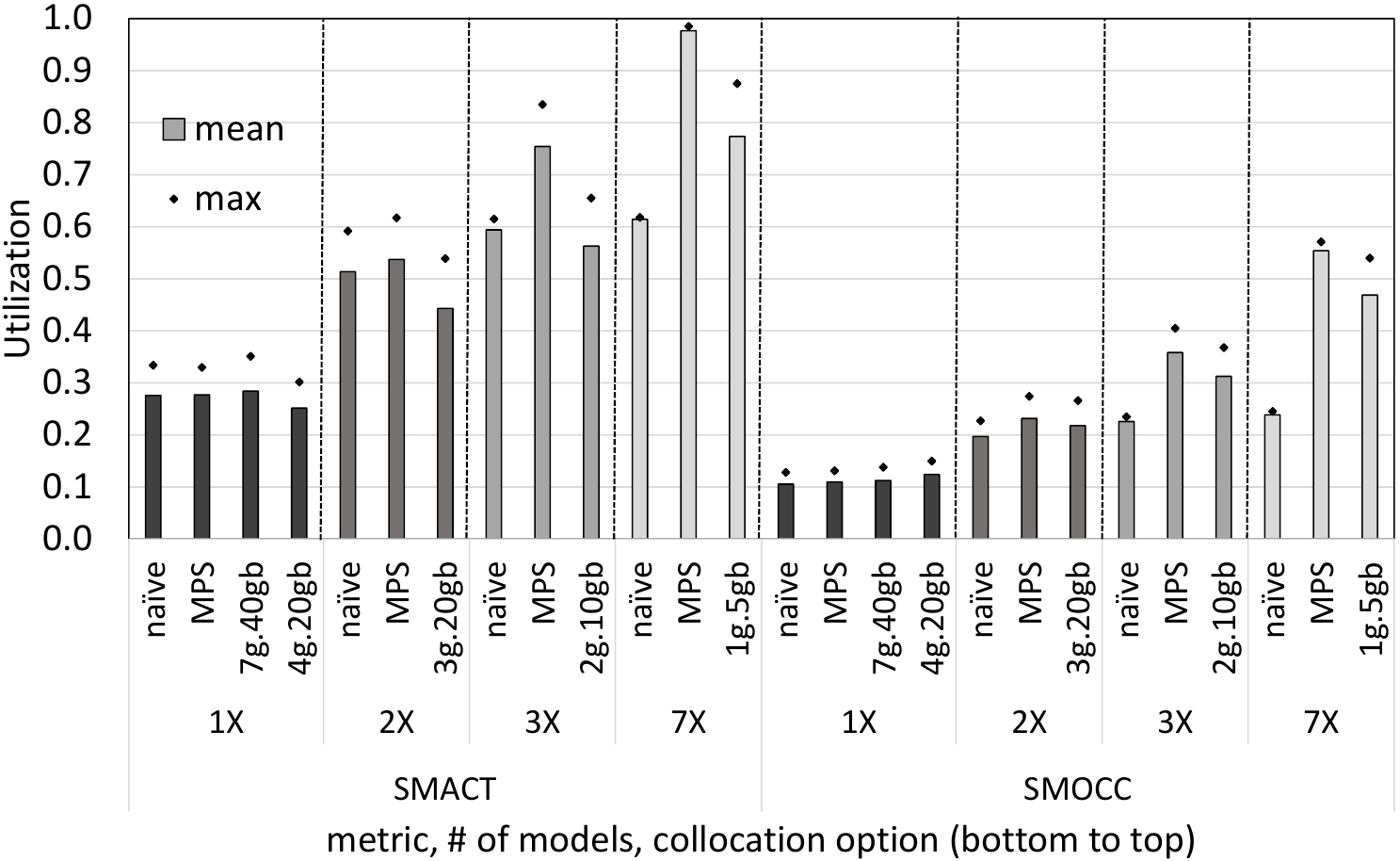}
\vspace{-5mm}
  \caption{GPU utilization}
  \label{fig:gract-resnet32-medium}
\end{subfigure}
\begin{subfigure}{.329\textwidth}
  \centering
\includegraphics[width=\linewidth]{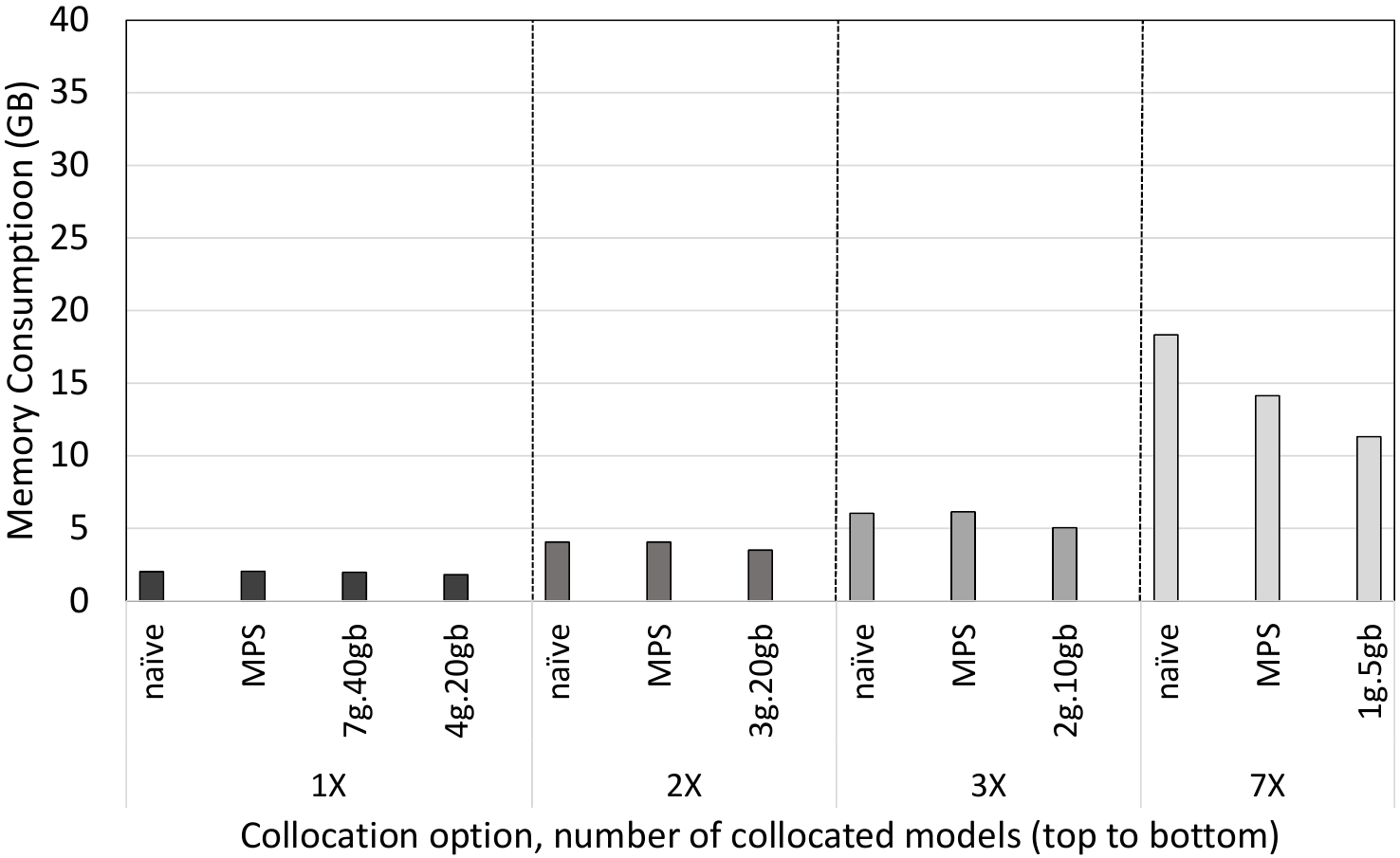}
\vspace{-5mm}
  \caption{Memory footprint}
  \label{fig:drama-resnet32-medium}
\end{subfigure}
\vspace{-4mm}
\caption{Medium: ResNet50 + ImageNet64 (batch size = 32).}
\label{fig:medium-resnet32-all}
\end{figure*}

\section{Results}
\label{sec:results}
Figures~\ref{fig:small-resnet32-all}-\ref{fig:large-cait-all} illustrate the results for
our uniform collocation experiments.
Each figure shows the results of a particular model and dataset combination (as listed in \Cref{table:datasets}).
Bars that are grouped together form one collocated workload with models trained in parallel.
The different degrees of collocation are seperated by dotted vertical lines.
The four non-collocated cases, which do not run any models in parallel, are the first four bars and form our baselines.
We omit the figures for two of the large model and dataset combinations:
(1) Resnet152 with batch size 128 and ImageNet and (2) EfficientNet with ImageNet.
These two cases were too large to allow for any collocation;
they ran out of memory on the GPU for all collocation mechanisms even when training just two models in parallel.
Sections~\ref{sec:results-ttc}-\ref{sec:results-gpum} present the results each focusing on a metric we collect (\Cref{sec:metrics})
for Figures~\ref{fig:small-resnet32-all}-\ref{fig:large-cait-all}.

After presenting the results for the uniform collocation,
\Cref{sec:results-hetero} describes our findings on the collocation runs using mixed-set of vision models,
and \Cref{sec:results-recommender} presents the effectiveness of collocation when training a recommender and a vision model in collocated fashion.

\subsection{Time per Epoch}
\label{sec:results-ttc}
As mentioned in \Cref{sec:metrics},
time per epoch is our main performance metric when comparing the effectiveness of different collocation methods.
The rest of the metrics are used to explain certain trends in the time per epoch results.

\begin{figure*}[!ht]
\centering
\begin{subfigure}{.329\textwidth}
  \centering
\includegraphics[width=\linewidth]{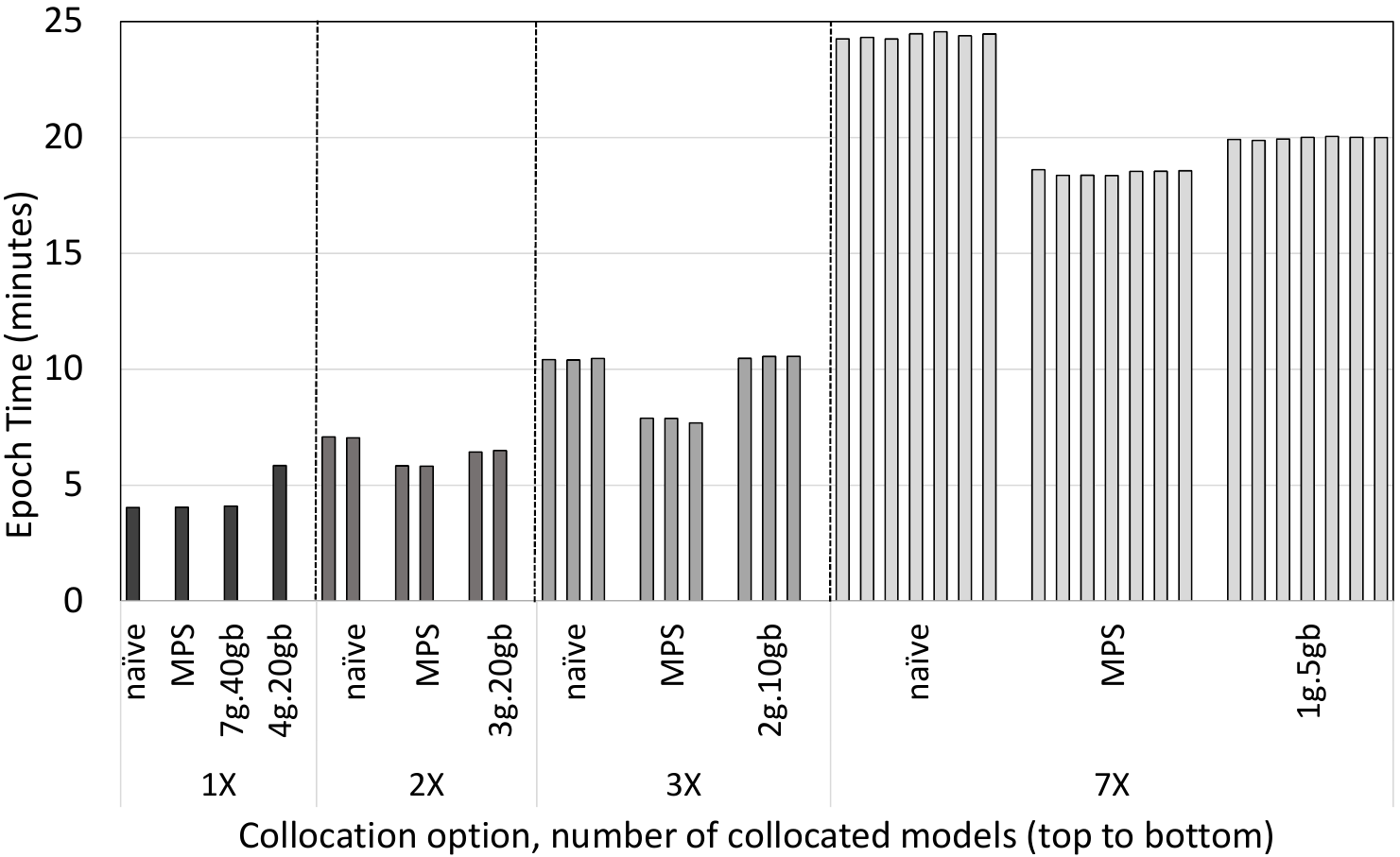}
\vspace{-5mm}
  \caption{Epoch time}
  \label{fig:time-resnet128-medium}
\end{subfigure}
\begin{subfigure}{.329\textwidth}
  \centering
\includegraphics[width=\linewidth]{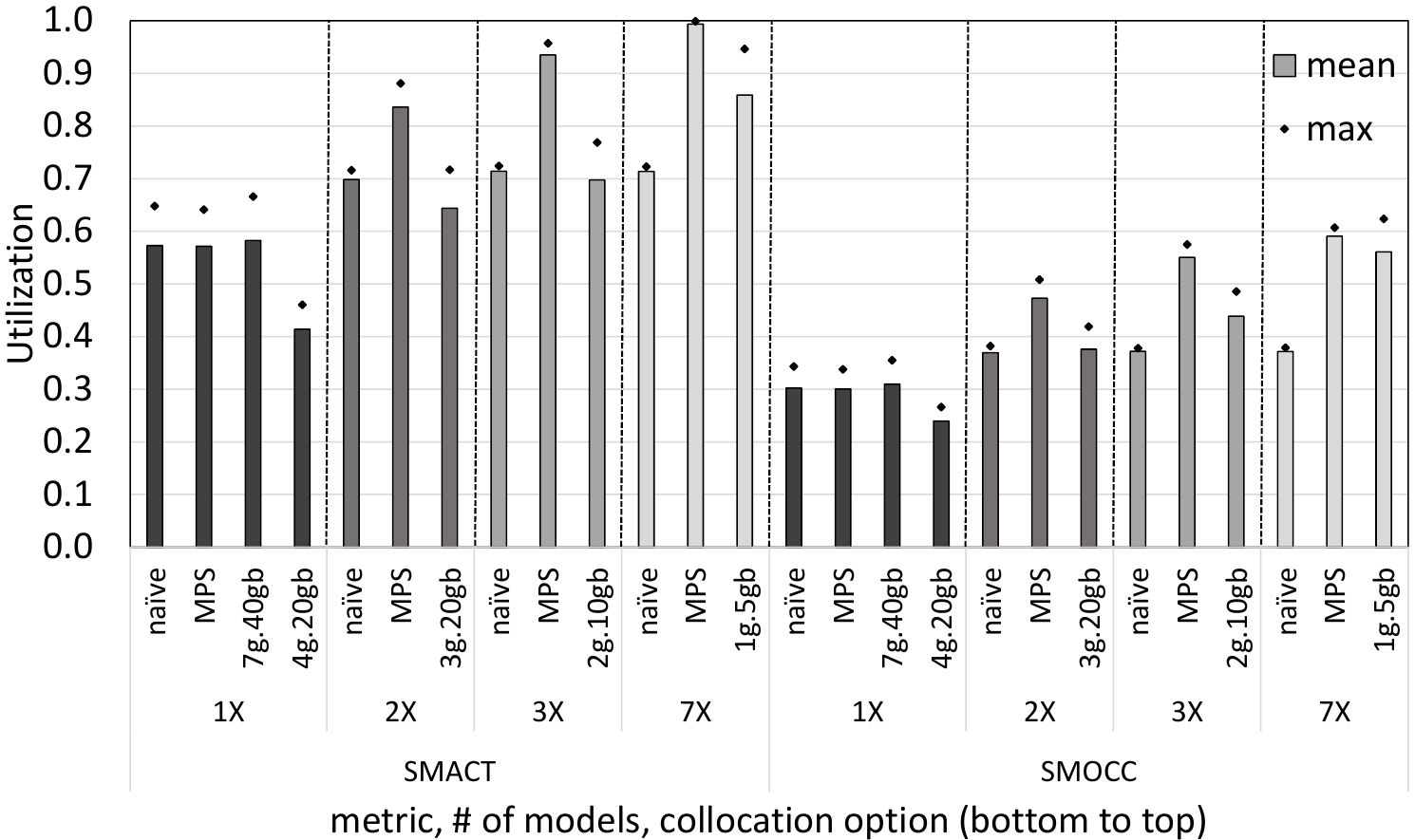}
\vspace{-5mm}
  \caption{GPU utilization}
  \label{fig:gract-resnet128-medium}
\end{subfigure}
\begin{subfigure}{.329\textwidth}
  \centering
\includegraphics[width=\linewidth]{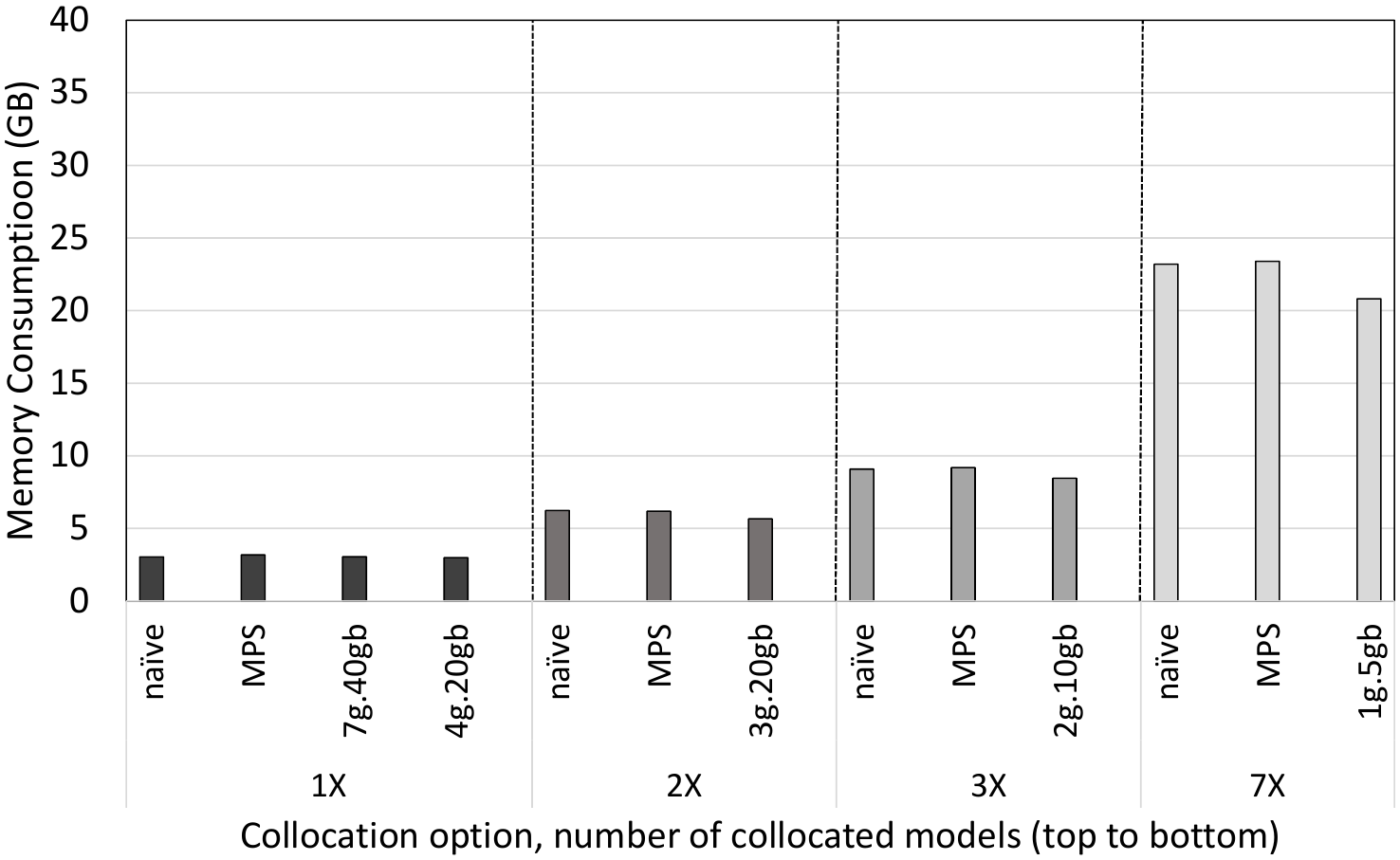}
\vspace{-5mm}
  \caption{Memory footprint}
  \label{fig:drama-resnet128-medium}
\end{subfigure}
\vspace{-4mm}
\caption{Medium: ResNet50 + ImageNet64 (batch size = 128).}
\label{fig:medium-resnet128-all}
\end{figure*}

\begin{figure*}[!ht]
\centering
\begin{subfigure}{.329\textwidth}
  \centering
\includegraphics[width=\linewidth]{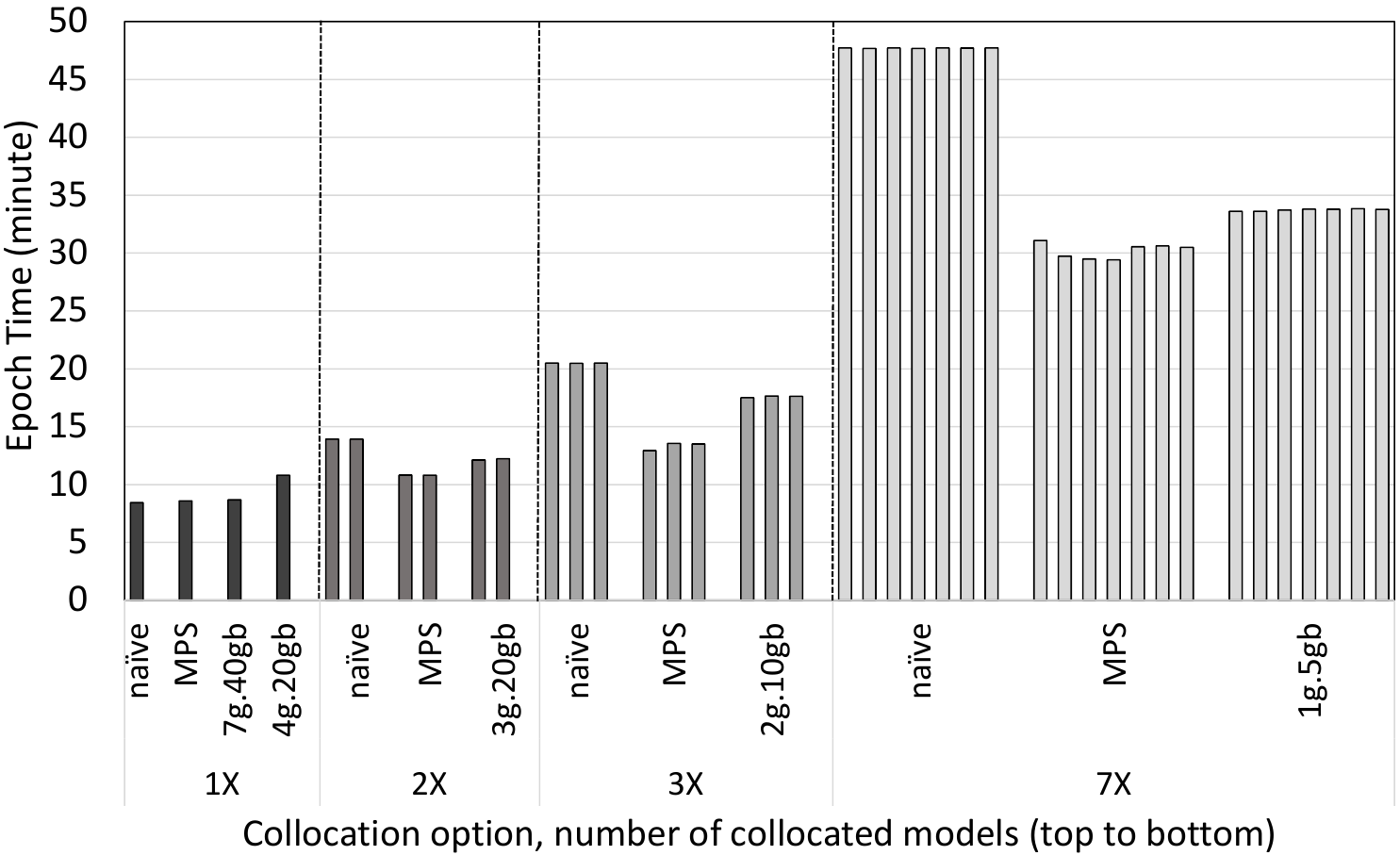}
\vspace{-5mm}
  \caption{Epoch time}
  \label{fig:time-efnet-medium}
\end{subfigure}
\begin{subfigure}{.329\textwidth}
  \centering
\includegraphics[width=\linewidth]{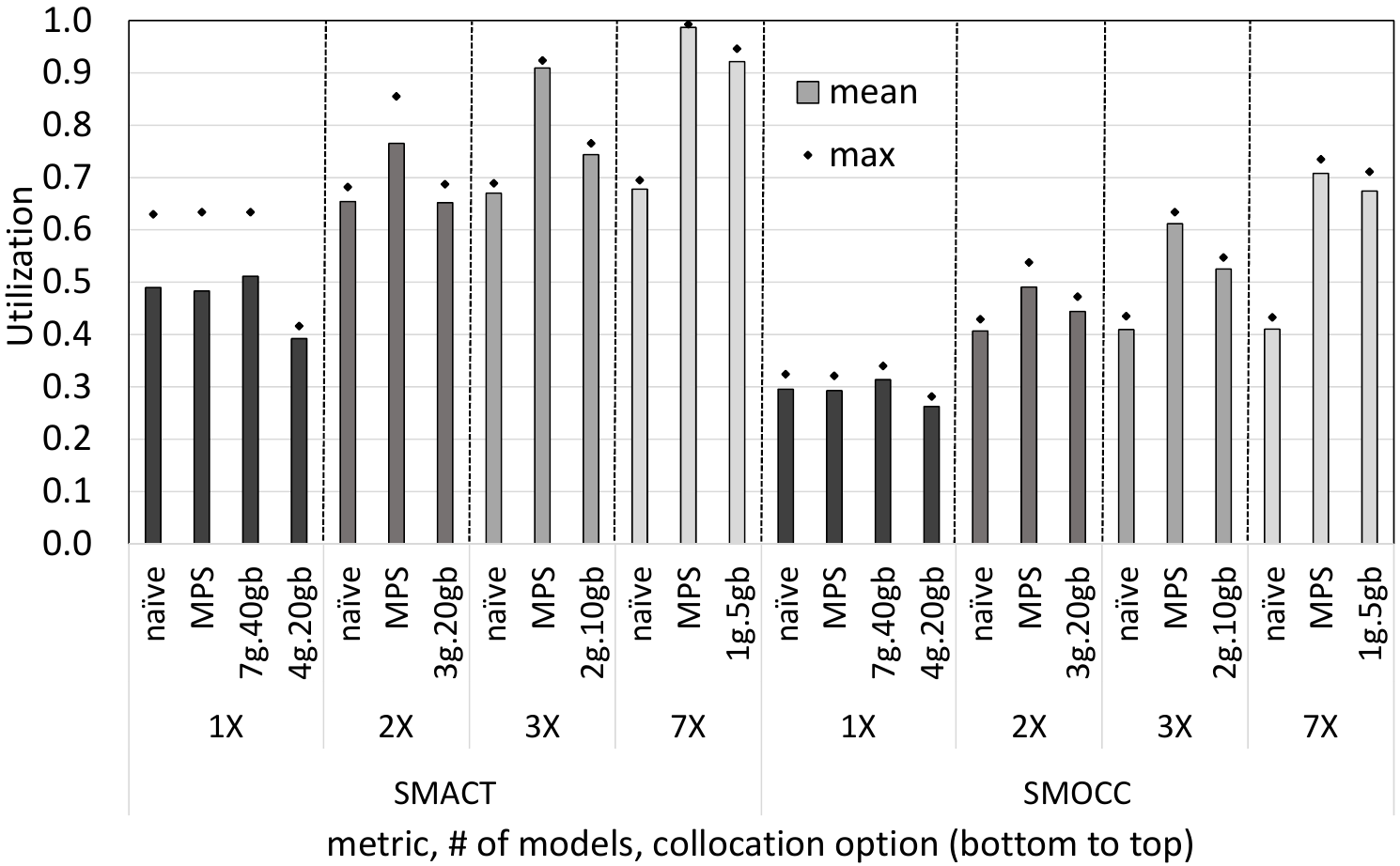}
\vspace{-5mm}
  \caption{GPU utilization}
  \label{fig:gract-efnet-medium}
\end{subfigure}
\begin{subfigure}{.329\textwidth}
  \centering
\includegraphics[width=\linewidth]{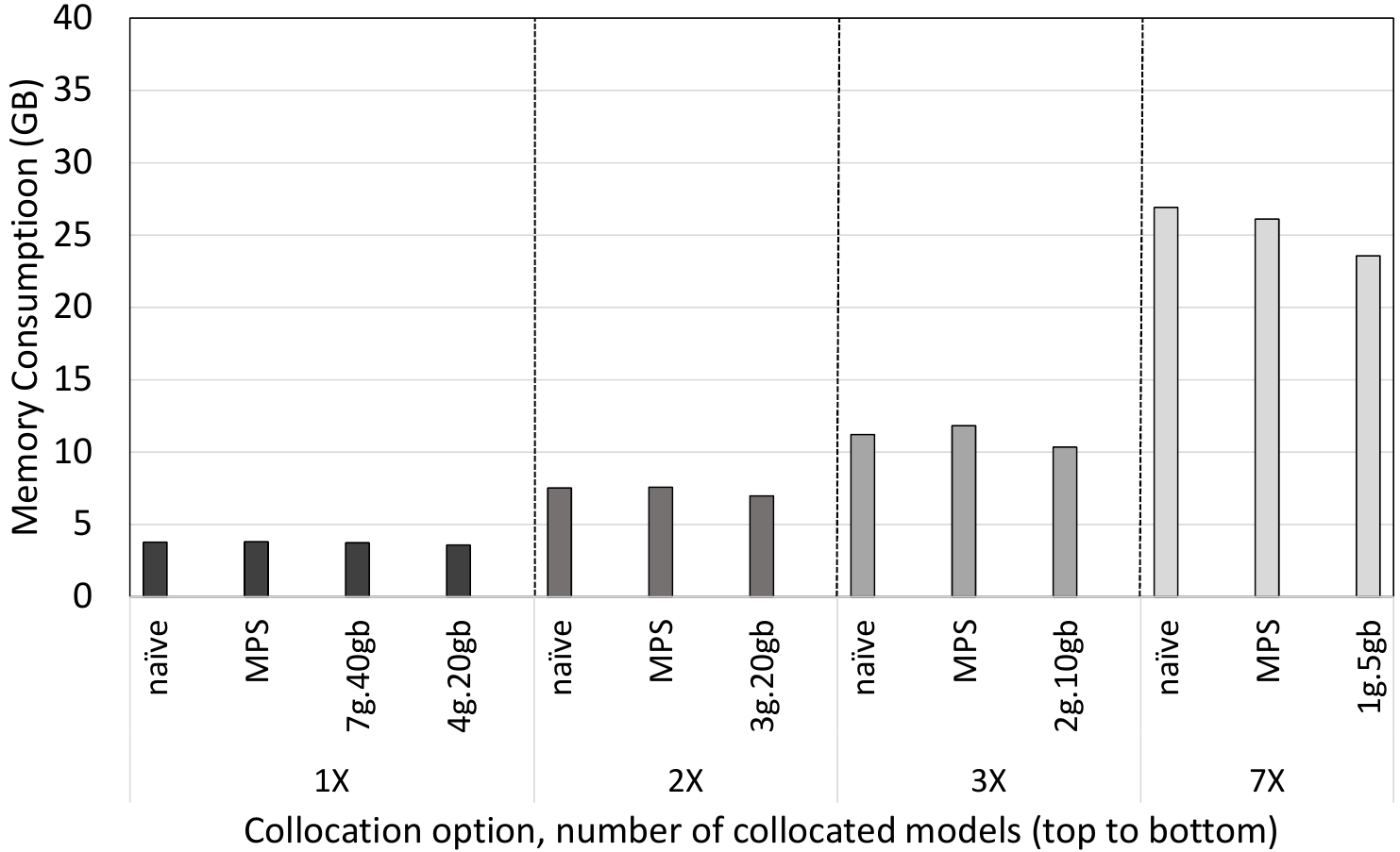}
\vspace{-5mm}
  \caption{Memory footprint}
  \label{fig:drama-efnet-medium}
\end{subfigure}
\vspace{-4mm}
\caption{Medium: EfficientNet\_s + ImageNet64 (batch size = 128).}
\label{fig:medium-efnet-all}
\end{figure*}

Starting with our baselines, we need to verify whether enabling the MPS daemon or MIG partitions introduce any visible overheads.
Looking at the first four bars of Figures~\ref{fig:time-resnet32-small}-\ref{fig:time-cait},
reveals that there is a little variation between the first three non-collocated workloads:
\texttt{naïve}, \texttt{mps}, and \texttt{7g.40gb}.
This indicates that there is negligible overhead of having MPS or enabling MIG over the naïve case.
On the other hand,
we see the impact of having fewer resources available on the \texttt{4g.20gb} MIG instance as the workloads get larger.
In Figures~\ref{fig:time-resnet128-medium}-\ref{fig:time-cait},
the single training run exhibits a larger time per epoch on the \texttt{4g.20gb} instance
compared to the other non-collocated runs.


Going over to the collocated runs,
comparing across the different collocation mechanisms on Figures~\ref{fig:time-resnet32-small}-\ref{fig:time-cait} reveals that MIG-based collocation performs better as the degree of parallelism increases (especially to 7).
MPS reveals itself as a clear winner, offering the best performance across the board.
In contrast, naïve collocation is the least effective. 

For the small ResNet models and 7-way collocation, the benefits of collocation become very visible.
With ResNet's embarrassingly parallel nature and the larger batch size allowing even more parallelism,
they manage a high utilization of the GPU compute resources without overloading the GPU, as demonstrated by \Cref{fig:gract-resnet128-small}.
The medium ResNet models reflect the same pattern, though start hitting compute resource boundaries under 7-way collocation, as seen in \Cref{fig:gract-resnet128-medium}. 
As a result, collocation provides considerable benefits for the small and medium cases with MIG and especially with MPS.

As expected, collocation impacts the time it takes to train the individual models.
Aside from 2-way collocation with our smallest workload (\Cref{fig:small-resnet32-all}), 
training more models in parallel increases the time to finish training a model.
Additionally, as the degree of collocation increases, so does the total time to train the models.
On the other hand, multiple models finish training simultaneously, increasing training throughput.
For example, except for the large workloads, 
2-way collocation delivers two models in roughly the same time as no-collocation delivers one model.
3-way collocation with MPS and MIG leads to a 15-35\%, 50-160\%, and 45-110\% increase in time per epoch compared to non-collocated case
for ResNet with batch size 32 (Figures~\ref{fig:time-resnet32-small} \& \ref{fig:time-resnet32-medium}),
ResNet with batch size 128 (Figures~\ref{fig:time-resnet128-small} \& \ref{fig:time-resnet128-medium}),
and EfficientNet (Figures~\ref{fig:time-efnet-small} \& \ref{fig:time-efnet-medium}), respectively,
while delivering three model training runs instead of one.
7-way collocation with MPS and MIG only increases the runtime by 40-80\% for our smallest workload
(\Cref{fig:small-resnet32-all}) while delivering 7 models in parallel.
These results clearly show that collocation is valuable when a single training run is not large enough for the available GPU compute and memory resources; e.g., the \textit{small} and \textit{medium} cases.

However, the picture shifts considerably with the large workloads (Figures~\ref{fig:large-resnet32-all} \& \ref{fig:large-cait-all}).
We no longer see considerable improvements for all of the collocated runs. MPS remains strong and is the only form of collocation that remains beneficial even at 3-way collocation.
Under naïve collocation, one epoch of training takes roughly as long as training the models in sequence without collocation. MIG fairs little better under 2-way collocation, but is not advantageous. Additionally, 7-way collocation becomes impossible due to memory constraints. 

\textit{\textbf{Take-away.} Collocation provides a significant increase in throughput, especially for smaller models. MPS collocation consistently beats the other forms of collocation, with MIG just slightly behind for workloads with smaller models.}

\subsection{GPU Utilization}
\label{sec:results-gract}
\begin{figure*}[!ht]
\centering
\begin{subfigure}{.329\textwidth}
  \centering
\includegraphics[width=\linewidth]{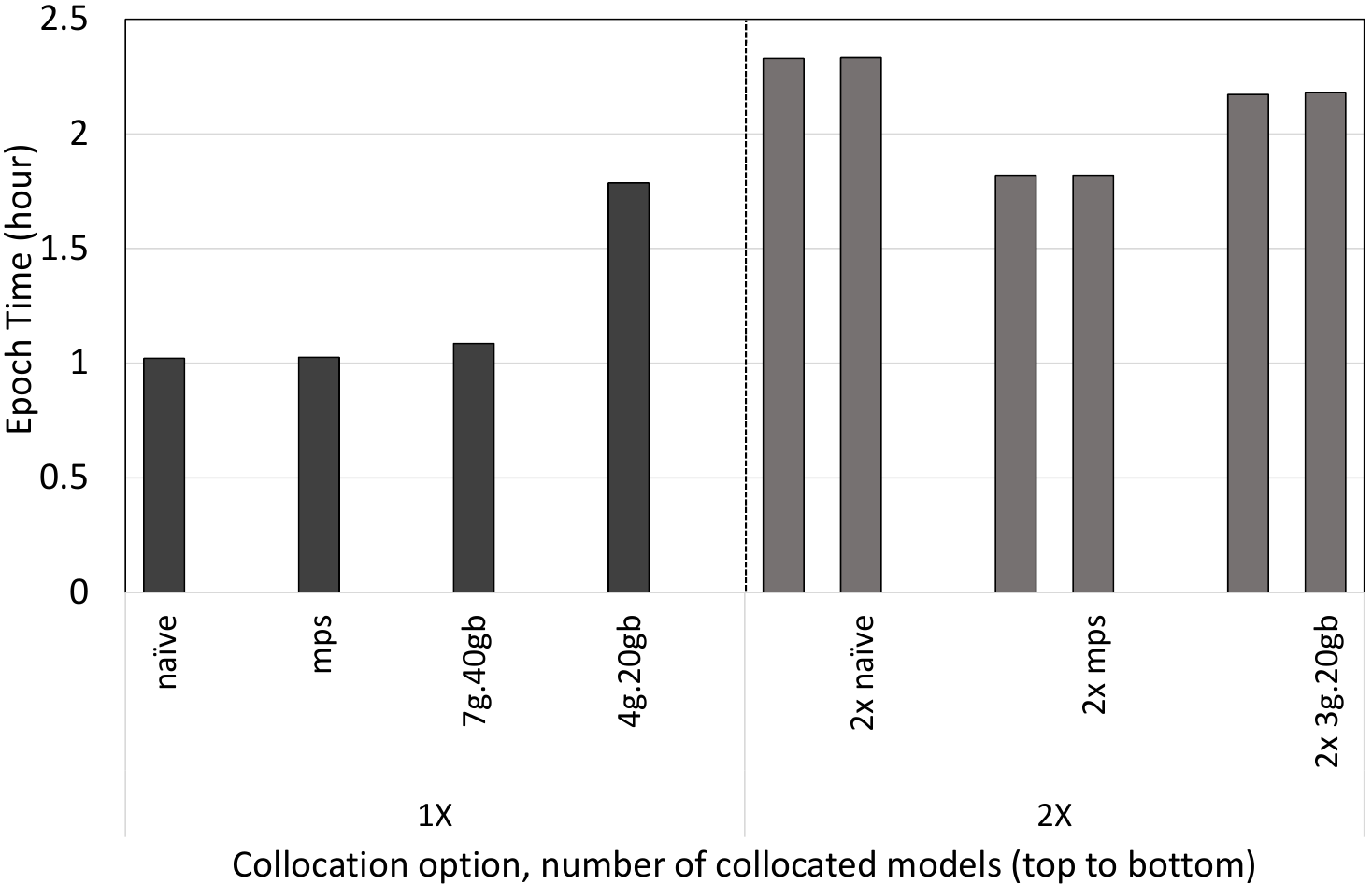}
\vspace{-5mm}
  \caption{Epoch time}
  \label{fig:time-cait}
\end{subfigure}
\begin{subfigure}{.329\textwidth}
  \centering
\includegraphics[width=\linewidth]{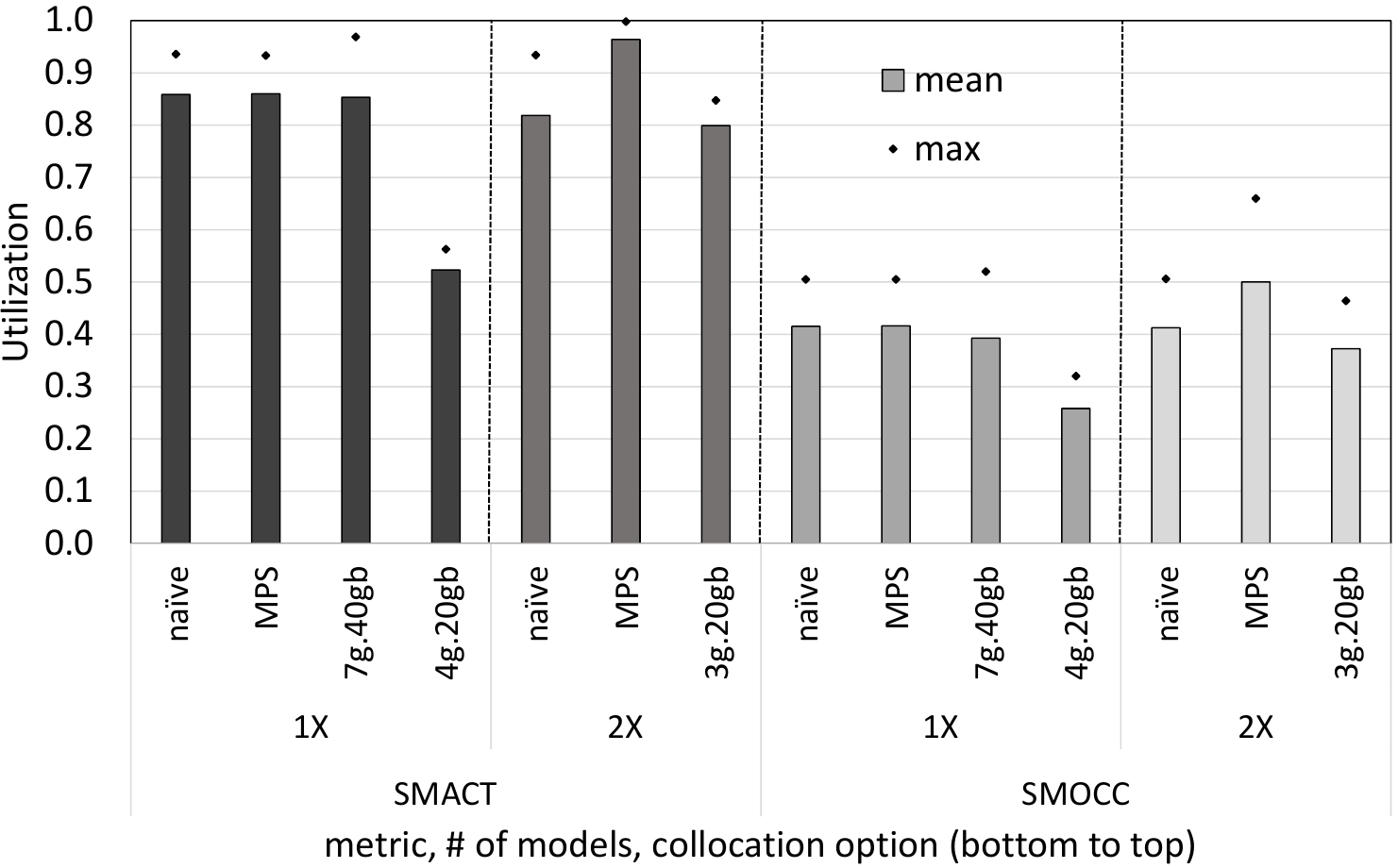}
\vspace{-5mm}
  \caption{GPU utilization}
  \label{fig:gract-cait}
\end{subfigure}
\begin{subfigure}{.329\textwidth}
  \centering
\includegraphics[width=\linewidth]{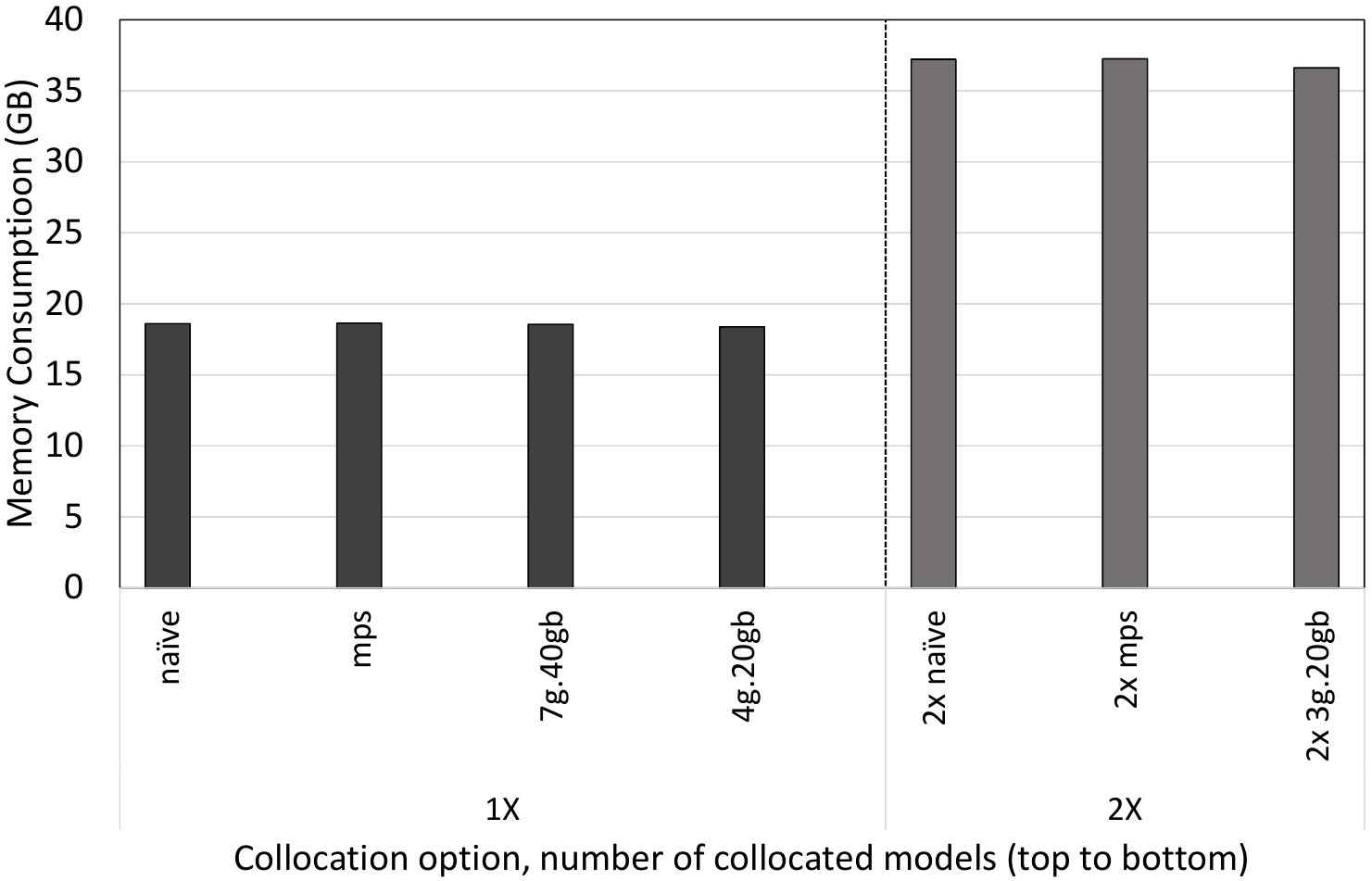}
\vspace{-5mm}
  \caption{Memory footprint}
  \label{fig:drama-cait}
\end{subfigure}
\vspace{-4mm}
\caption{Large: CaiT + ImageNet (batch size = 128).}
\label{fig:large-cait-all}
\end{figure*}

Observing GPU utilization in Figures~\ref{fig:gract-resnet32-small}-\ref{fig:gract-cait}
helps us understand when collocation provides benefits and when it does not.
In general,
the lower the GPU utilization for the non-collocated scenarios,
the higher the benefits of collocation.

As explained in \Cref{sec:metrics},
the utilization \% reported for MIG is with respect to the aggregate available compute resources (98 SMs) to MIG.
A small amount of compute resources of the GPU is lost to MIG overhead.
Results for naïve and MPS depict the utilization of the whole GPU (108 SMs).


\begin{figure*}[!ht]
\centering
\begin{subfigure}{.329\textwidth}
  \centering
\includegraphics[width=\linewidth]{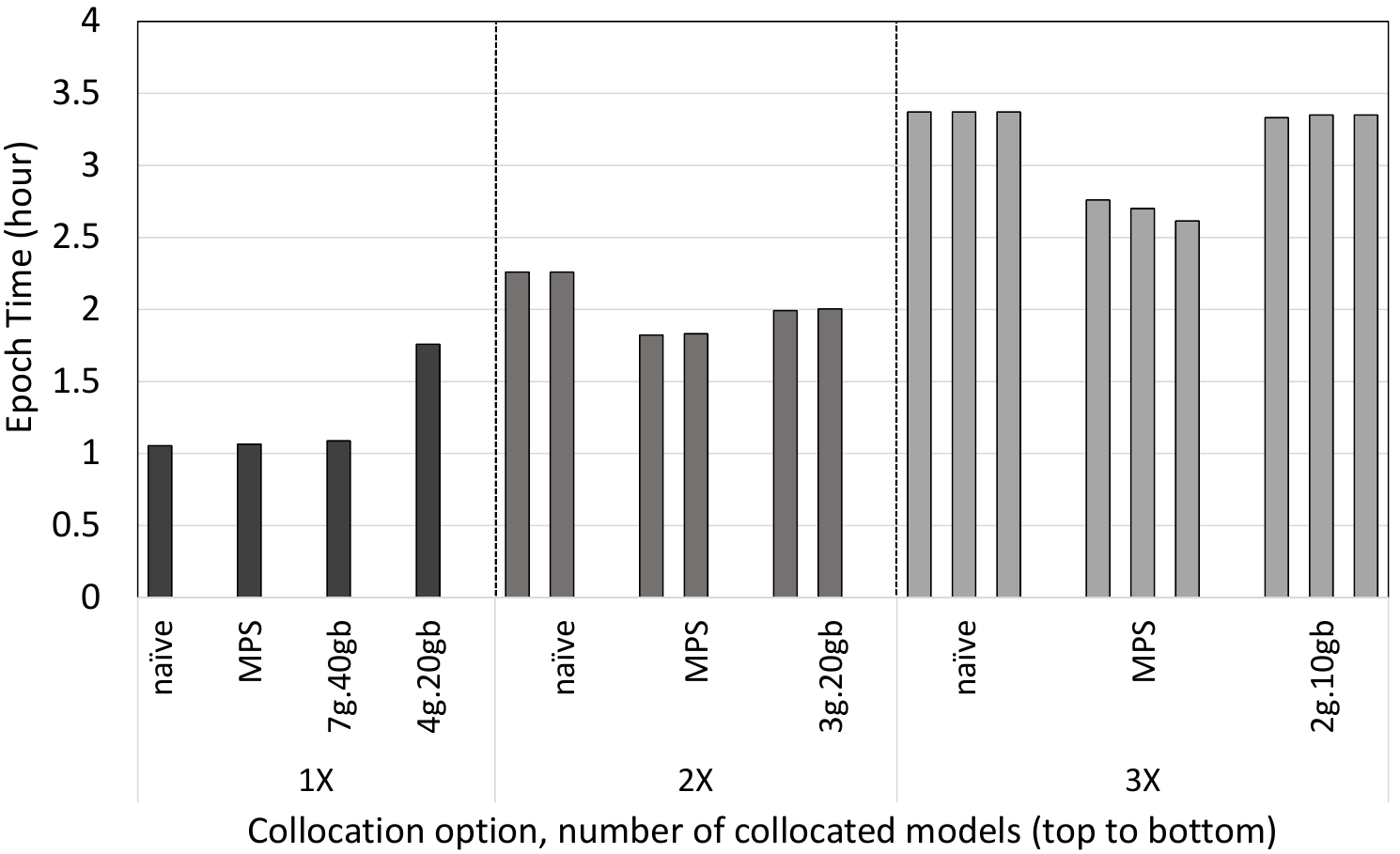}
\vspace{-5mm}
  \caption{Epoch time}
  \label{fig:time-resnet32-large}
\end{subfigure}
\begin{subfigure}{.329\textwidth}
  \centering
\includegraphics[width=\linewidth]{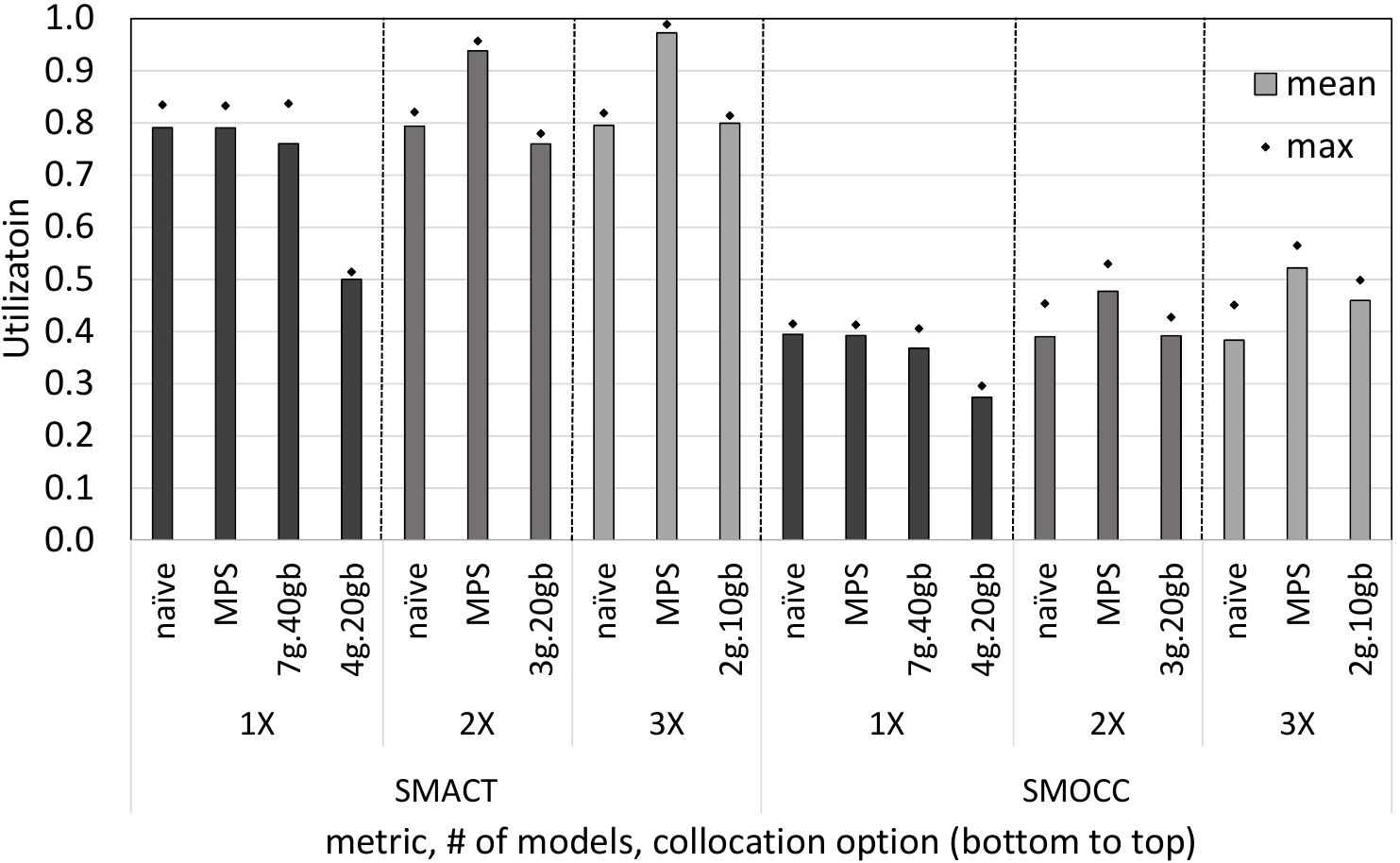}
\vspace{-5mm}
  \caption{GPU utilization}
  \label{fig:gract-resnet32-large}
\end{subfigure}
\begin{subfigure}{.329\textwidth}
  \centering
\includegraphics[width=\linewidth]{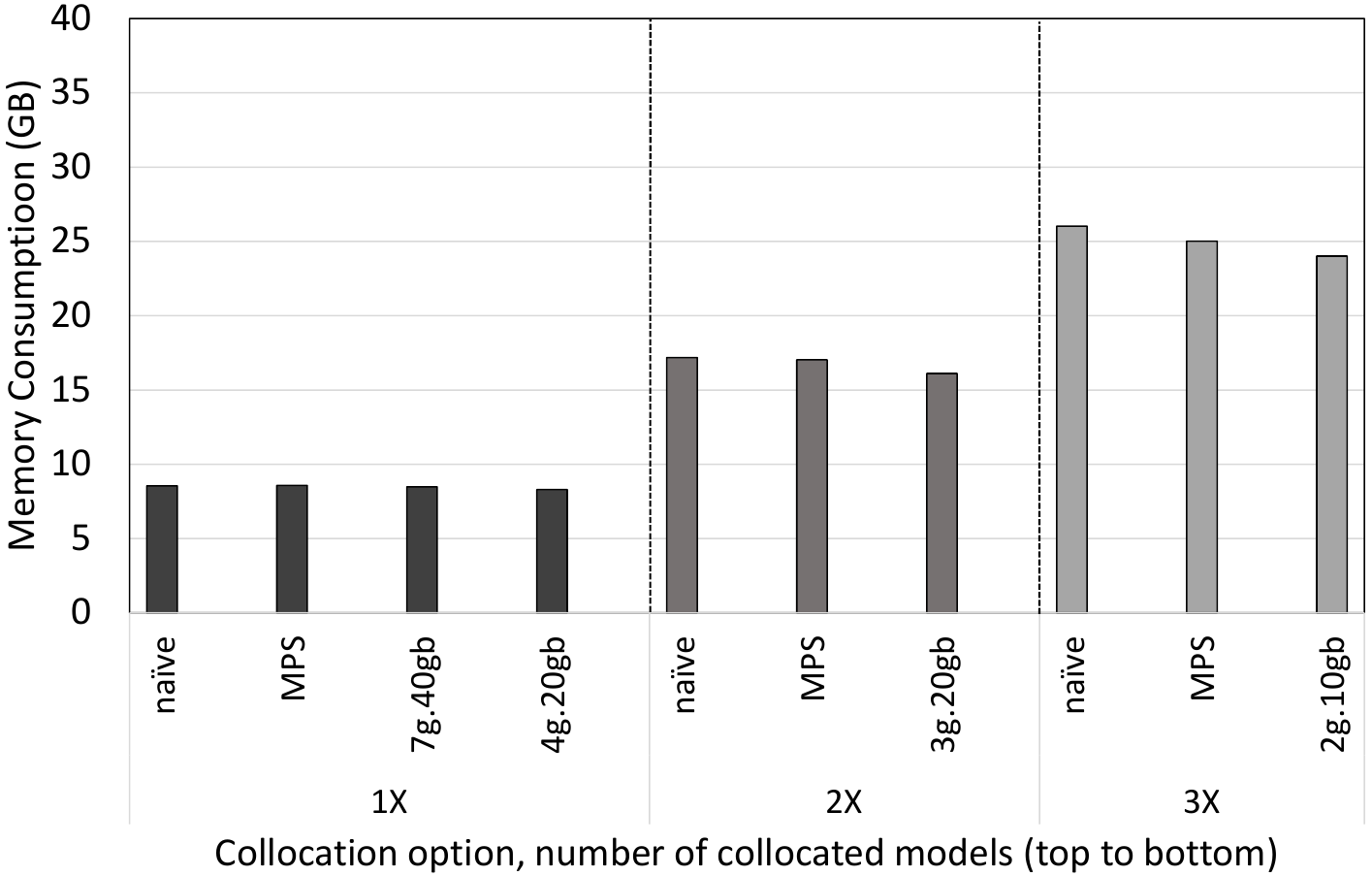}
\vspace{-5mm}
  \caption{Memory footprint}
  \label{fig:drama-resnet32-large}
\end{subfigure}
\vspace{-4mm}
\caption{Large: ResNet152 + ImageNet (batch size = 32).}
\label{fig:large-resnet32-all}
\end{figure*}


Starting with the small workloads (ResNet26 with batch size 32 in \Cref{fig:gract-resnet32-small}),
we can see that non-collocated runs do not fully utilize the streaming multiprocessors.
Collocation massively increases the utilization, allowing for more of the GPU to be useful when training 2, 3, or 7 models at the same time. Both SM activity and occupation do not meet the saturation point for this small use-case, explaining the excellent collocation performance as discussed in \Cref{sec:results-ttc}.

Most of the patterns seen in the small case are present in the medium ResNet case (\Cref{fig:gract-resnet32-medium}) as well.
The utilization is notably higher, with MPS 7-way collocation (\texttt{1g.5gb}) approaching maximal SM activity, though occupation does not max out.
The results for large ResNets (\Cref{fig:gract-resnet32-large}) notably deviates from the these two experiments.
The non-collocated runs already have notably high utilization and the 2-way collocated runs get close to the limit. Going one step further, the 3-way collocated runs over-saturate the compute side of the GPU
leading to diminishing returns in terms of the throughput achieved under collocation (\Cref{sec:results-ttc}).

As we increase the batch size from 32 to 128 for ResNet,
GPU utilization jumps by up to 75\% and 110\% in, respectively, the
small (\Cref{fig:gract-resnet128-small}) and medium (\Cref{fig:gract-resnet128-medium}) cases.
The GPU utilization under collocation with MPS is especially high, reaching almost 100\% SM activity, with MIG reaching similar numbers on 7-way collocation in the medium case. 

As in the case of time per epoch results,
the GPU utilization of EfficientNet (Figures~\ref{fig:gract-efnet-small} \& \ref{fig:gract-efnet-medium})
is similar to the results of ResNet with batch size 32.
There is a big jump in utilization from non-collocation to collocation.
MIG and MPS collocation feature high utilization, showing very similar numbers for 7-way collocation, with MPS having a small utilization lead on MIG for 2- and 3-way collocation.

Finally, for CaiT (\Cref{fig:gract-cait}), there is little variety in the GPU utilization across different cases.
Even though the utilization numbers are very similar, MPS manages to perform significantly better than both naïve and MIG collocation. It is the only form of collocation that provides a throughput benefit in this case over training the models in series.

\textit{\textbf{Take-away.}
MPS consistently provides the highest GPU utilization, regardless of workload.
Additionally, throughput under MPS fares better despite heavy GPU utilization.}


\subsection{Memory Footprint}
\label{sec:results-gpum}
The GPU memory footprint of the models while training is crucial as it is the main determinant for whether models can be trained in collocated fashion.
As we have seen in the previous two sections,
when there is high utilization of compute resources on a GPU,
the collocated runs can still make forward progress, even though the collocation might not be beneficial in terms of throughput.
On the other hand, 
when there is not enough memory available for the aggregate memory footprint of the collocated training runs,
then these models run out of memory when assigned to the GPU.

Figures~\ref{fig:drama-resnet32-small}-\ref{fig:drama-cait}
report the aggregate memory footprint on the GPU for different collocation methods for each workload. 
They demonstrate that the increase in memory footprint with collocation is proportional to the degree of collocation.
Notably, MIG collocation shows slightly smaller memory footprints than the two other options.
In general, 2-way, 3-way, and 7-way collocation result in roughly two, three, and seven times the memory footprint of no collocation, respectively.
This is an expected result as the models are not sharing data across collocated runs in the scope of this study, even though all the collocated runs use the same dataset.

The memory footprint for MIG prompted us to delve deeper into PyTorch's memory allocation.
The reduced memory allocation for MIG shows up in both nvidia-smi readings and PyTorch's advanced memory statistics.
However, PyTorch's basic memory allocation and reservation trackers, which count space allocated for the tensors and reserved by the allocator, respectively, do not show any difference across the different collocation methods.
Training the models on a separate GPU with less available GPU memory displays the same pattern of slightly reduced memory footprint, confirming that this is not a symptom unique to MIG, but is rather due to the memory size available to PyTorch.
In other words, PyTorch adjusts the memory footprint depending on the total available memory, which is less in the case of non-\verb|7g.40gb| MIG instances compared to whole GPU memory available under MPS and naïve. 
Switching the memory allocator backend from PyTorch's native implementation to CUDA's built-in asynchronous allocator removes the differences in the memory footprint of different collocation methods. 
However, we do not recommend this switch as it slows down the training process.

\textit{\textbf{Take-away.} Memory requirements for uniformly collocated models can be effectively estimated by multiplying the memory required by a single model. PyTorch allocates slightly less memory under MIG instances that have a fraction of the whole memory than under other collocation methods.}

\subsection{Mixed Vision Workloads}
\label{sec:results-hetero}
The results presented so far focused on homogeneous collocation scenarios.
Such cases can be extremely useful in practice when a data scientist is performing hyper-parameter tuning to come up with the ideal set of parameters for a model
repeatedly running the same model with a different set of parameters. 
On the other hand, 
there is also value to investigate non-homogeneous collocation scenarios
to observe what happens when individual training runs stress the GPU unequally.

Based on how models of different sizes behaved in our previous experiments, 
we select combinations of small, medium, and large ResNet models with corresponding dataset sizes to collocate for the heterogeneous runs (as listed in \Cref{table:mixedvision}).
We opted to keep a static MIG configuration while testing heterogeneous collocation
since in a real-world scenario, e.g., in a data center, the MIG partitions would already be set
and re-partitioning after each training run could be impractical.

\begin{figure}[t]
\centering
\begin{subfigure}{.31\textwidth}
  \centering
\includegraphics[width=\linewidth]{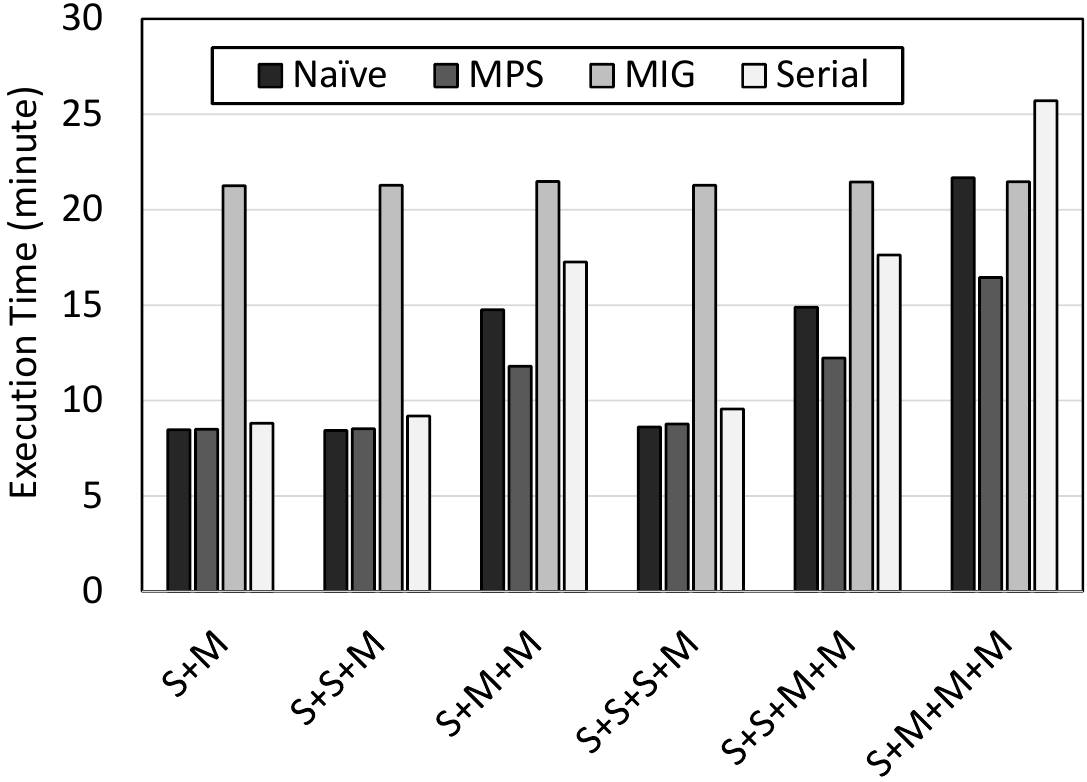}
  \label{fig:mixed_bars_m}
\end{subfigure}
\begin{subfigure}{.16\textwidth}
  \centering
\includegraphics[width=\linewidth]{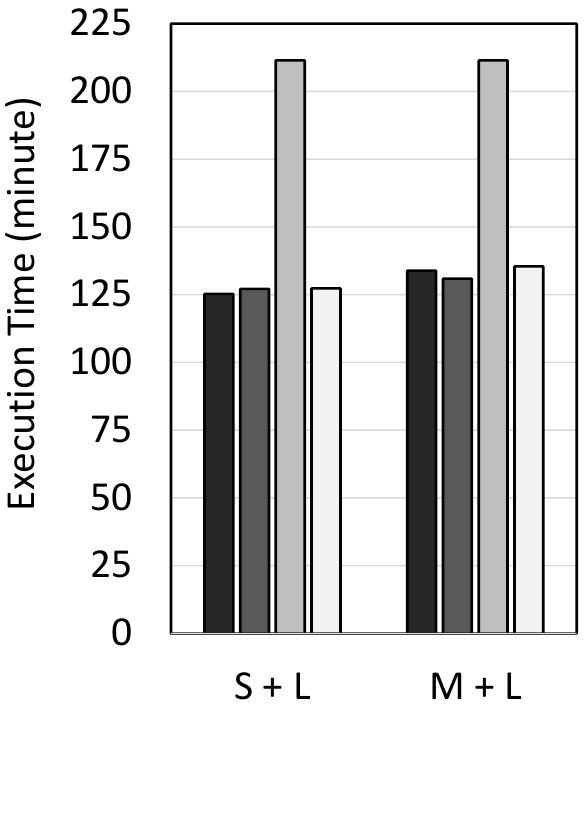}
  \label{fig:mixed_bars_l}
\end{subfigure}
\vspace{-6mm}
\caption{Total time for training mixed vision workloads with (naïve, MPS, MIG) \& without (serial) collocation for two epochs. Workload configurations can be found in \Cref{table:mixedvision}.
}
\label{fig:mixed_all_exec}
\end{figure}

\Cref{fig:mixed_all_exec} details the total execution time for training the collocated models using the different collocation methods in comparison to training them back to back, \textit{serial}, without collocation.
We see that the benefits of collocation vary heavily across workloads.
For small workloads such as "S+M" and "S+S+M",
naïve and MPS collocation provide sizeable benefits
by training the small model without impacting the medium one. In general, MPS collocation provides the largest benefit, and its flexibility over MIG is a great advantage here.

MIG performs significantly worse for these kinds of workloads as the resources available to the medium model are limited even after the small model finishes. 
On the other hand,
when increasing the workload by including more medium-sized models, e.g. "S+M+M+M",
this disadvantage of MIG diminishes, allowing it to slightly surpass naïve collocation.
This is in line with conclusions in \Cref{sec:results-ttc}.
Note that there is little difference in the execution time achieved by MIG
in each of the graphs in \Cref{fig:mixed_all_exec}.
MIG isolates the collocated models,
and the resources available to these medium models are identical under MIG.
Thus, with MIG, the total execution time boils down to the execution time of the slowest model training regardless of the mix of the collocation. 

As \Cref{sec:results-gract} discussed, the large model runs utilize more of the GPU on their own.
"S+L" and "M+L" runs corroborate this earlier finding as collocation offers reduced benefit in these cases.
MIG performs significantly worse than the other options as the resource availability of the large model is locked even after the small and medium models finish their execution.

\begin{figure}[t]
\centering
\begin{subfigure}{.48\textwidth}
  \centering
\includegraphics[width=\linewidth]{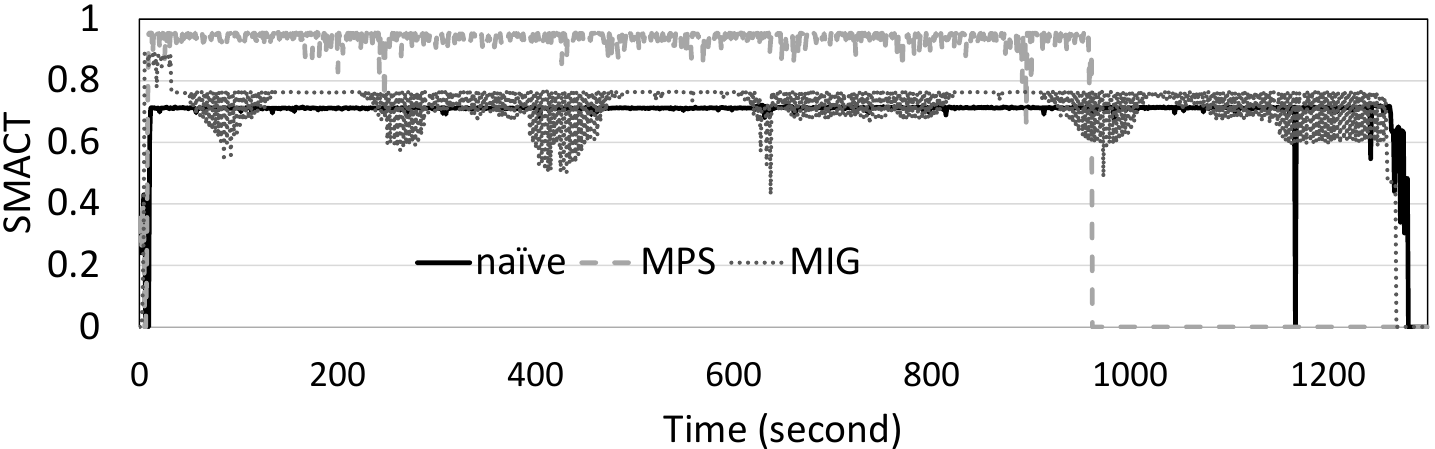}
\vspace{-5mm}
  \caption{Streaming Multiprocessor Activity (SMACT)}  \label{fig:mixed_util_mss}
\end{subfigure}
\vspace{-2mm}
\begin{subfigure}{.48\textwidth}
  \centering
\includegraphics[width=\linewidth]{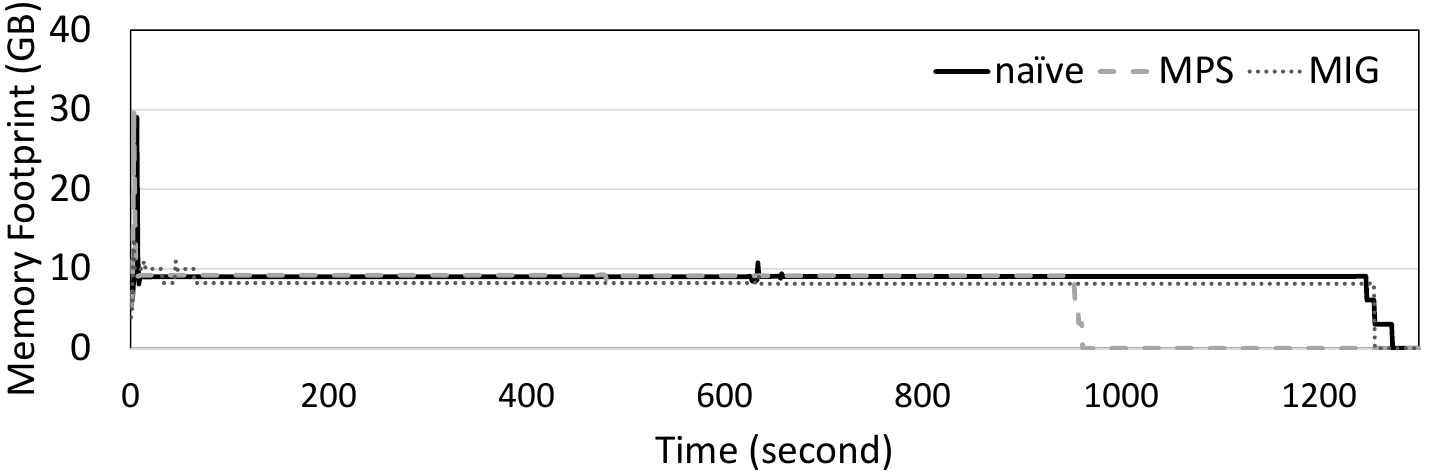}
 \centering
  \caption{Memory footprint}
  \label{fig:mixed_mem_mss}
\end{subfigure}
\caption{GPU utilization and memory footprint over time for S+M+M+M from \Cref{fig:mixed_all_exec}.}
\label{fig:mixed_all_gpu}
\end{figure}

\Cref{fig:mixed_all_gpu} dives deeper into the "S+M+M+M" workload to observe how the GPU utilization and memory footprint changes over time during collocated runs with naïve, MPS, and MIG collocation.
We pick this mix as it is the one that utilizes MIG instances the best.
The GPU utilization under MIG gets lowered after the small model finishes,
since MIG is unable to fill-up the corresponding instance with more work.
On the other hand, naïve and MPS are able to keep similar GPU utilization throughout.
In contrast, the memory footprint follows a similar trend for all collocation strategies. 
It is higher in the beginning as all four models are training.
The values then drop off quickly once the small model finishes training.

\textit{\textbf{Take-away.} MPS collocation provides significant benefits when training a mix of models. MIG collocation is unsuitable when mixing different models as resources can not be re-allocated once some of the models finish training.}

\subsection{Mixed Recommender and Vision Workloads}
\label{sec:results-recommender}

\Cref{table:mixedrec} shows the results for collocating a recommender model with a vision model. As the recommender model takes much longer to train, we measure its training time by iterations instead. One training block contains 102400 training iterations and a validation pass. The ResNet training time is measured per epoch as in previous experiments. We treat the first training block and epoch as warm-up to ensure correct measurements.
In this particular case, the MIG-based collocation runs on compute instances that share memory as the recommender model does not fit into the memory of any of the smaller GPU instances.

\begin{table}[t]
\centering
\caption{Mixed collocation with Recommender model. Recommender time is for one training block (102400 iterations) plus validation. ResNet time is for one epoch. The reported increase in time (\%) is relative to the two no-collocation runs at the top.}
\resizebox{8.1 cm}{!} {
\begin{tabular}{|c|c|c|c|c|}
\hline
Workload     & Recom. Time (h)  & ResNet Time (h)  & SMACT & Memory (GB)\\ \hline \hline
Recommender  & 5.36         & -            & 5\%   & 29.14  \\ \hline
ResNet152    & -            & 1.05         & 82\%  & 8.47   \\ \hline \hline
Naive        & 6.09 (+14\%) & 1.11 (+5\%)  & 81\%  & 37.75  \\ \hline
MPS          & 5.57 (+4\%)  & 1.10 (+4\%)  & 81\%  & 37.62  \\ \hline
MIG (shared) & 5.60 (+5\%)  & 1.40 (+33\%) & 39\%  & 37.86  \\ \hline
\end{tabular}
}
\label{table:mixedrec}
\end{table}

Adding a memory-heavy model such as the recommender greatly promotes collocation. While training time does increase when collocating these models, it only goes up between 4\%-14\% and 4\%-33\% for the recommender and ResNet, respectively. MPS performs especially well, training both models with just a 4\% increase in training time. Interestingly, the availability of extra memory under a shared memory MIG configuration benefits the training of ResNet, noting improved performance over \texttt{4g.20gb} in Figure~\ref{fig:time-resnet32-large} even though the model is collocated with the recommender.

As before, memory consumption roughly corresponds to the sum of both models. SM activity, however, does not increase on collocation.
This suggest that the slowdown under collocation is due to another resource contention. Under MIG, only part of the compute power of the GPU can be assigned to ResNet, even though the recommender requires little. 

\textit{\textbf{Take-away.} Collocation of models that stress different parts of the GPU may greatly increase throughput in exchange for a minimal increase in training time.}

%% file: sections/5-summary.tex
\section{Guidelines \& Challenges}
\label{sec:summary}
Based on the results of our experiments in \Cref{sec:results},
we now provide some guidelines for deep learning training collocation in \Cref{sec:guidelines},
and highlight the challenges faced when doing performance analysis in this domain in \Cref{sec:challenges}.

\subsection{Collocation Guidelines}
\label{sec:guidelines}
\textit{\textbf{Workload collocation is highly beneficial when the aggregate compute and memory needs of the collocated deep learning training runs fit the GPU.}}
In \Cref{sec:results},
we observe significant throughput benefits (up to four times) for small compute-intensive workload setups
despite the increase in the epoch time of individual training runs.
Medium-sized compute-intensive workloads also exhibit similar throughput benefits, though less pronounced.
Similarly, collocating compute- and memory-intensive training together leads to a more effective use of the hardware resources without significantly hindering training time.

\textit{\textbf{Collocation gives diminishing returns when the SM activity of an individual training run is already close to 100\%.}}
For example, even without collocation, the training of CAIT hits SMACT numbers of 90\% and up, significantly higher than any other model.
This shows that the GPU's compute resources are utilized almost completely.
Hence, the large workload scenarios, like CAIT, do not benefit as much from collocation.
A user can make an educated guess for the most effective degree of collocation (no collocation, 2-way, 3-way, etc.) based on the SMACT values of an individual run.

\textit{\textbf{The aggregate memory footprint of the collocated runs can simply be estimated by the sum of the memory footprints of the individual runs and cannot exceed the available memory on the GPU.}}
As a result, for large workload scenarios, we either cannot collocate any training runs (e.g., ResNet152 with batch size 128) or cannot reach beyond 2- or 3-way collocation (e.g., Figures \ref{fig:large-resnet32-all} \& \ref{fig:large-cait-all}).
A user can determine whether a set of runs can be collocated effectively a priori based on the known or expected memory footprints of individual runs.

\textit{\textbf{MPS achieves better performance across the board thanks to its flexible distribution of hardware resources among the collocated runs.}}
Hence, for setups where just a single user is submitting training jobs, MPS-based collocation will always be the better option over naïve and MIG.

\textit{\textbf{MIG's rigid partitioning leads to sub-optimal performance compared to MPS but is able to support collocation when a strict separation is required among the runs.}}
When there are multiple users submitting training jobs or when even a single user requires non-interfering runs due to e.g. privacy concerns, MIG is the only option for collocation.
If the workload is known a priori, the ideal set of MIG instances can be created accordingly.
This way, MIG-based collocation can still be beneficial
over running the training runs serially
even though it comes at a slight cost in performance compared to MPS.

%% file: sections/6-discussion.tex
\subsection{Challenges}
\label{sec:challenges}

Benchmarking in a rapidly evolving field like deep learning,
has its challenges, which we encountered in our study.

\textbf{Maturity of the toolset for MIG.}
DCGM is capable of tracking metrics per MIG instance as mentioned in \Cref{sec:metrics}.
However, earlier in our study, it didn't reliably report the metrics for the \verb|4g.20gb| instance. 
This issue has since been resolved, allowing us to include \verb|4g.20gb| in \Cref{sec:results}.
On the other hand, there are other reporting anomalies under MIG with DCGM.
For example, metrics that track the data movement across PCIe,
which connects the CPU and the GPU, 
do not report anything when MIG instances are used.

\textbf{PyTorch improvements.}
When we started our study, we had CUDA 11.6 and PyTorch 1.13 as the latest versions.
Under this configuration, MIG exhibited more competitive results, beating both naïve and MPS in 7-way collocated uniform workloads.
With the recent updates for both CUDA and PyTorch,
we repeated our experiments with CUDA 11.7 and PyTorch 2.0.
The results in \Cref{sec:results} are with these latest versions.
With this new configuration, MPS became significantly faster, reducing epoch time previously exhibited by MPS by up to a factor of three.
With additional experiments, we verified that the improved performance of MPS was mainly due to the PyTorch version update rather than the CUDA driver updates.


%

%% file: sections/7-conclusion.tex
\section{Conclusion}\label{sec:conclusion}
In this paper, 
we did a performance characterization on a modern GPU device that has support for multiple means of GPU collocation: naïve, MPS, and MIG.

Our results demonstrate that GPU collocation is most useful for small and medium-sized workloads that cannot fully saturate the whole GPU.
Although per-model training is overall slower,
more work can be done per unit of time by executing workloads in parallel,
which utilizes the GPU resources more effectively and increases the training throughput.
MIG notably requires a rigid setup.
The GPU has to be partitioned beforehand and one has to know what instance sizes
are the most beneficial.
If the workload across the instances are imbalanced,
runs that finish early will leave some instances idle, unless
there is other work that could be allocated over those instances.
Naïve collocation and MPS, on the other hand, can utilize the resources released by the finished work, increasing the training performance of models that require more time to train.
In general, MPS provides the best collocation performance.

In this work, 
we limited our focus to training on a single GPU, since NVIDIA doesn't allow multi-GPU training with MIG.
In a data center,
many workloads can be collocated not only on the same GPU but also on the same server.
Therefore, studying the impact of collocation
while running other workloads on other GPUs on the same device would be interesting future work.
Furthermore,
considering the results with the recommender model, further
investigations of the shared memory instances of MIG could be worthwhile.